\documentclass{article}

\usepackage{microtype}
\usepackage{graphicx}
\usepackage{subcaption}
\usepackage{booktabs} 

\usepackage{hyperref}

\usepackage[ruled,vlined]{algorithm2e}

\usepackage[preprint]{icml2026}

\usepackage{amsmath}
\usepackage{amssymb}
\usepackage{mathtools}
\usepackage{amsthm}

\usepackage{mathrsfs} 
\usepackage{float}
\usepackage{multirow}

\usepackage[capitalize,noabbrev]{cleveref}

\theoremstyle{plain}
\newtheorem{theorem}{Theorem}[section]
\newtheorem{proposition}[theorem]{Proposition}

\theoremstyle{definition}

\theoremstyle{remark}

\usepackage[textsize=tiny]{todonotes}

\icmltitlerunning{ }

\begin{document}

\twocolumn[
  \icmltitle{Preserving Spectral Structure and Statistics in Diffusion Models}



  \icmlsetsymbol{equal}{*}

  \begin{icmlauthorlist}
    \icmlauthor{Baohua Yan}{cu}
    \icmlauthor{Jennifer Kava}{cu}
    \icmlauthor{Qingyuan Liu}{cu}
    \icmlauthor{Xuan Di}{cu}
  \end{icmlauthorlist}

  \icmlaffiliation{cu}{Department of Civil Engineering and Engineering Mechanics, Columbia University, NY, United States}

  \icmlcorrespondingauthor{sharon.di}{sharon.di@columbia.edu}

  \icmlkeywords{Machine Learning, ICML}

  \vskip 0.3in
]



\printAffiliationsAndNotice{}  

\begin{abstract}

    Standard diffusion models (DMs) rely on the total destruction of data into non-informative white noise,
    forcing the backward process to denoise from a fully unstructured noise state.
    While ensuring diversity, this results in a cumbersome and computationally intensive image generation task.
    We address this challenge by proposing new forward and backward process within a mathematically tractable spectral space.
    Unlike pixel-based DMs, our forward process converges towards an informative Gaussian prior $\mathcal{N}({\boldsymbol{\hat{\mu}}}, \boldsymbol{\hat{\Sigma}})$ rather than white noise.
    Our method, termed \textbf{Pre}serving Spectral \textbf{S}tructure and \textbf{S}tatistics (\textbf{PreSS}) in diffusion models, guides spectral components toward this informative prior while ensuring that corresponding structural signals remain intact at terminal time.
    This provides a principled starting point for the backward process, enabling high-quality image reconstruction that builds upon preserved spectral structure while maintaining high generative diversity.
    Experimental results on CIFAR-10, CelebA and CelebA-HQ demonstrate significant reductions in computational complexity, improved visual diversity, less drift, and a smoother diffusion process compared to pixel-based DMs.

\end{abstract}

\section{Introduction}
\label{sec:intro}
Diffusion models (DMs) have emerged as one of the most powerful classes of generative models, achieving remarkable success across diverse data modalities including image- \cite{ho2020denoising, DBLP:conf/iclr/0011SKKEP21, DBLP:conf/cvpr/RombachBLEO22, deepmind2025imagen4}, video-~\cite{DBLP:journals/corr/abs-2210-02303, DBLP:conf/cvpr/BlattmannRLD0FK23, videoworldsimulators2024, sora2_openai_2025} and audio-generation~\cite{DBLP:conf/iclr/KongPHZC21, DBLP:conf/iclr/Lam00022, DBLP:conf/icml/LiuCYMLM0P23}.
At the core of diffusion-based generative models~\cite{ho2020denoising} is a forward process that progressively corrupts clean data into a tractable isotropic Gaussian prior $\mathcal{N}(0, I)$, followed by a learned backward process that recovers the data distribution through iterative denoising. The remarkable success of these models is largely attributable to the framework that maps image data to white noise. Hence, this is a conventional mechanism predominantly incorporated in diffusion models and is widely regarded as a critical component to achieve high sample diversity and robust generative performance~\cite{DBLP:conf/cvpr/RombachBLEO22, DBLP:conf/cvpr/0010S00G24, kwon2022diffusion-f13, DBLP:conf/icml/NicholD21}.

\begin{figure}[t]
    \begin{center}
        \includegraphics[width=1.0\linewidth]{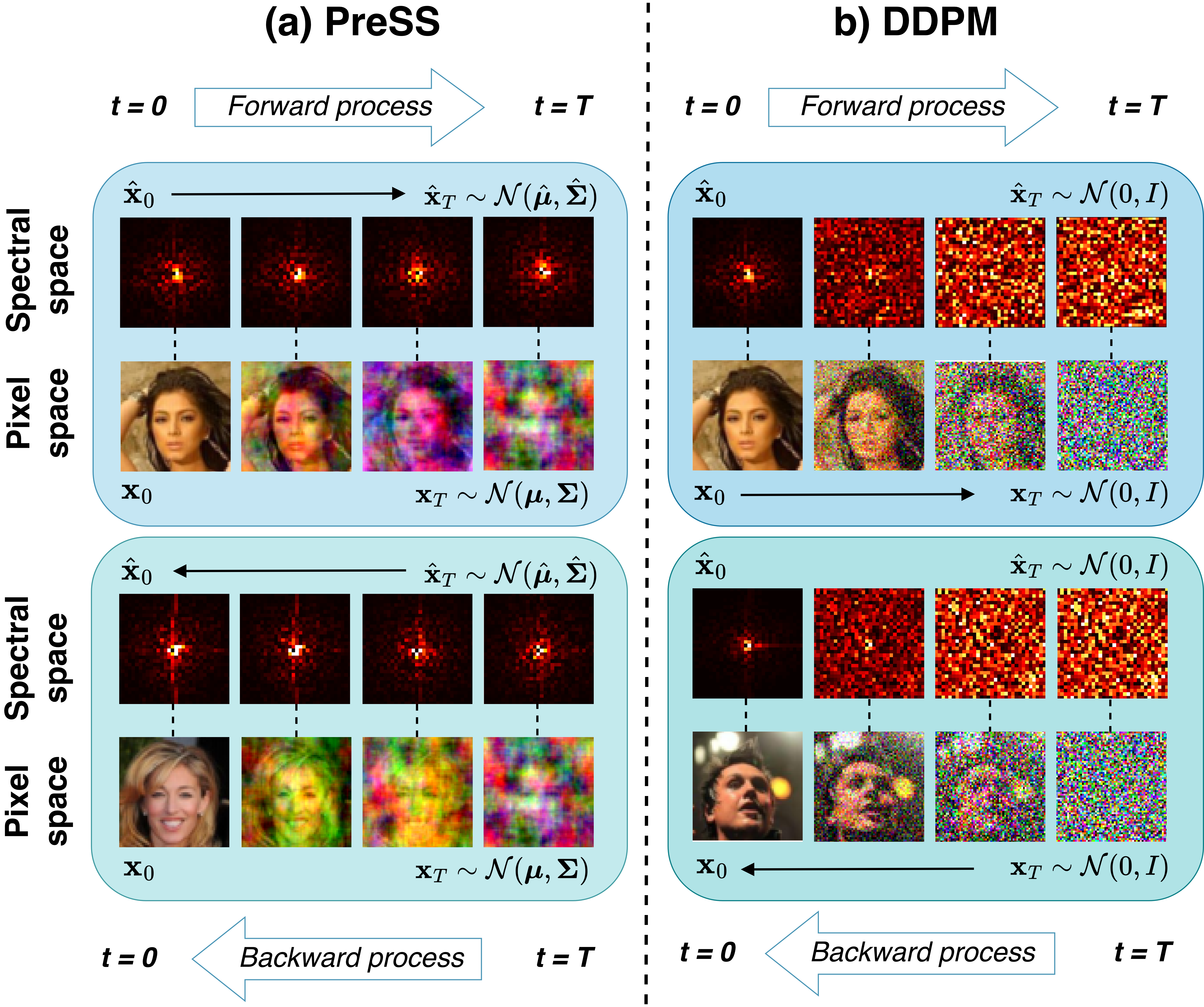}
    \end{center}
    \caption{\textbf{Visual comparison of (a) PreSS and (b) DDPM.} PreSS leverages preserved spectral components to synthesize more detailed and diverse samples (see Table~\ref{tab:quantitative_results_compact}), while DDPM follows the conventional noising to and denoising from white noise.}
    \label{fig:teaser_figure}
\end{figure}

While mathematically convenient, this framework treats all data components (pixels) as equally unimportant during the diffusing process and discards all useful information within image data, including geometric and statistical priors that uniformly appear among samples. Hence, data reconstruction heavily depends on the accuracy of a learned denoising function. While research work has been done on improving the U-Net architecture~\cite{DBLP:conf/icml/NicholD21} or by implementing other architecture replacements~\cite{DBLP:conf/iccv/PeeblesX23}, these approaches are generally time- and resource-intensive with a large degree of unpredictability, since the prediction of images or noise from nearly white noise is extremely difficult and intractable. This total destruction of data not only complicates image synthesis and leads to high sampling drift~\cite{DBLP:conf/nips/DarasDDD23} but also necessitates high computational costs and extensive sampling trajectories~\cite{DBLP:conf/iclr/SalimansH22} to recover a coherent image from an uninformative prior. This inherent limitation motivates the development of a model that preserves informative features throughout the diffusion process, enabling high-quality data generation while maintaining diversity.

\begin{figure}[h]
\begin{center}
\includegraphics[width=0.75\linewidth]{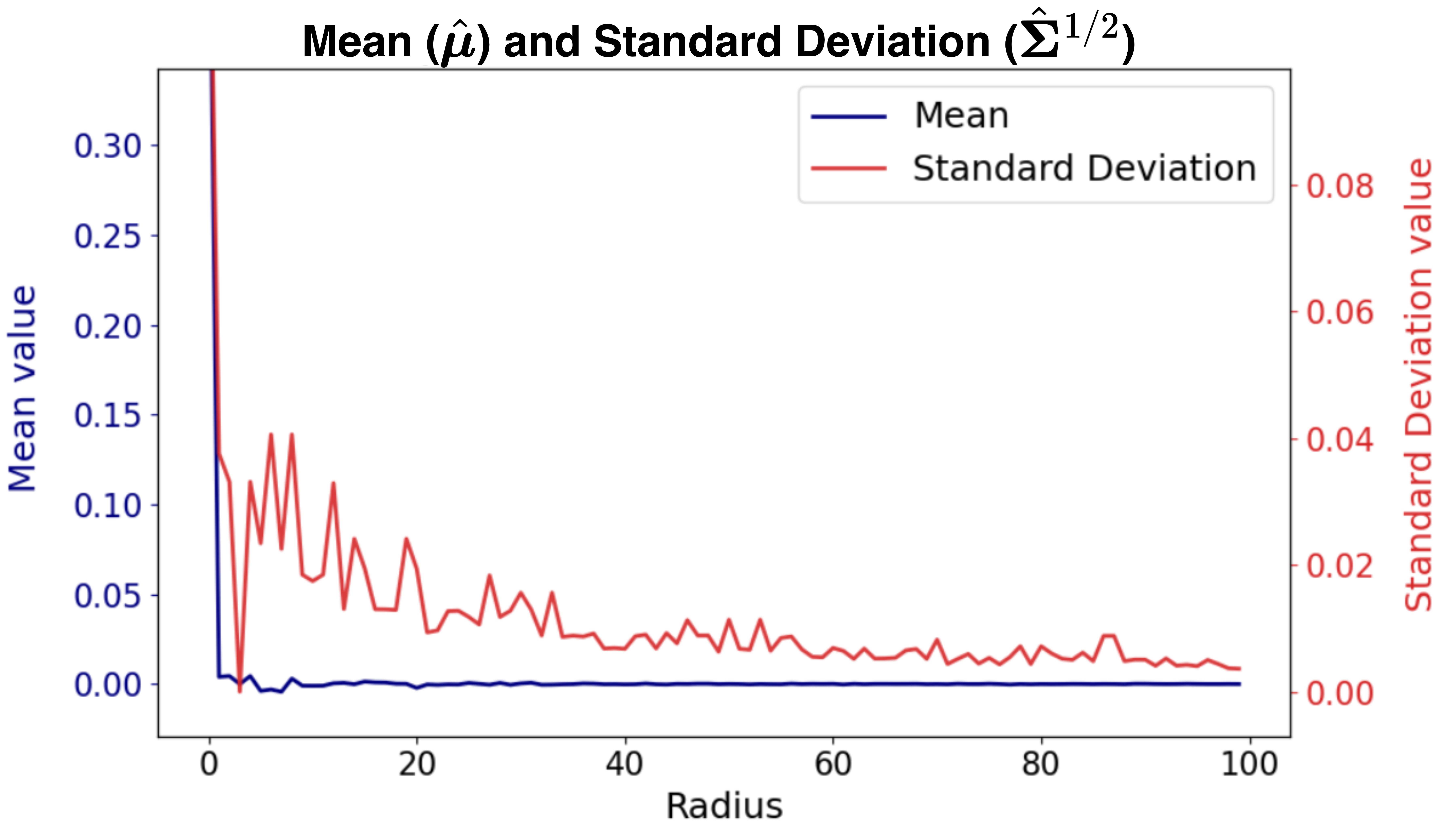}
\end{center}
\caption{Mean and standard deviation of the spectral components from the CIFAR-10 dataset. $x$-axis is the frequency radius of the spectral components. We pick red channel and show the curves. }
\label{fig:power-law}
\end{figure}

Due to the lack of statistical properties, it is difficult to further analyze preservable features from images in pixel space. However, we can turn to other mathematically interpretable latent spaces, such as spectral space to obtain a solution. By decomposing images via Fourier transform into spectral components, we are able to observe spectral structure and statistics that appear uniformly among image data. Notably, spectral components of natural images typically follow a $1/K^\alpha$ power-law distribution and are heavy-tailed in the spectral space~\cite{kolmogorov1991local, simoncelli2001natural, field1987relations, DBLP:conf/icml/NashMDB21}. Conventional image diffusion models also destroy this spectral structure into white noise and flatten the power-law distribution, erasing all structural inductive biases in the data~\cite{DBLP:journals/corr/abs-2502-10236}. In addition, we observe that the mean and variance of the spectral components also satisfy this power scaling behavior as illustrated in Figure~\ref{fig:power-law}. 

Motivated by the uniform existence of spectral structure and statistics, we propose a unified framework that \textbf{preserves spectral structures and statistics} as informative priors to guide the image reconstruction process, or \textbf{PreSS}, as we call it.
Our model comprises three core components:


(1) Fourier transform. Image data is converted via Fourier transform into spectral space where the entire diffusion process is carried out.
Importantly, performing the entire diffusion process in a mathematically interpretable spectral space enables the observation of statistical information and spectral features otherwise not observable in pixel space.
This inspires us to leverage spectral insights as informative priors to guide the diffusion process.

(2) Guided diffusion process. We propose a guided diffusion process that preserves spectral features by directing the forward process towards, and the backward process from, a target distribution $\mathcal{N}(\hat{\boldsymbol{\mu}}, \hat{\boldsymbol{\Sigma}})$ that respects the power-law distribution and statistics of spectral components.
By initializing the backward process with a statistically informed prior, we sidestep the need for the redundant and computationally intensive task of re-learning global and local structural information from scratch. This allows the model to allocate more capacity towards synthesizing fine details and textures while mitigating spectral drift, thus improving both generation quality and diversity.

(3) Incorporation into other DDPM-based models. By offering a principled alternative to the conventional DDPM, our framework can be easily applied to other DDPM-based models. This high degree of architectural compatibility allows PreSS to be incorporated into existing diffusion-based models with minimal modifications. It allows existing DDPM-based models to improve image generation quality and reduce computational complexity while demonstrating the potential generalizability of our models. Section~\ref{sec:extension} further discusses this approach. 


Theoretically, we establish that our model preserves spectral features corresponding to power-law distributed components. Empirically, we demonstrate that PreSS achieves significant improvements over DDPM on CIFAR-10, CelebA, and CelebA-HQ across multiple metrics. Beyond image synthesis, this method is naturally extensible to other data modalities with inherent spectral representations (e.g., audio, video, physical data and point clouds) where preserving inherent structural features are important to generate reliable and accurate samples. Overall, we hope that the concept of feature preservation will find broader applicability and our model can enable further development of generative models building upon DDPM.

\textbf{Related work.} Foundational diffusion frameworks, such as Denoising Diffusion Probabilistic Models (DDPMs) \cite{ho2020denoising} and score-based generative models \cite{DBLP:conf/iclr/0011SKKEP21} rely on a forward process that gradually corrupts data with isotropic Gaussian noise, followed by a learned backward process that iteratively denoises the data distribution into a generated image.
Despite achieving remarkable success in image synthesis, these models often suffer from significant computational overhead, which limits their scalability to image generation of higher resolutions.


Recent efforts have begun to examine diffusion processes in spectral space~\citep{DBLP:journals/corr/abs-2209-14125, DBLP:conf/icml/CrabbeHSS24, DBLP:journals/corr/abs-2410-02667}, consistently showing that the forward noising schedule governs how spectral components interact and evolve. Recent work~\cite{DBLP:journals/corr/abs-2502-10236} has investigated how shaping the forward process in the spectral space can better align with image statistics.
EqualSNR~\cite{DBLP:journals/corr/abs-2505-11278} observes that standard diffusion models corrupt different frequency bands at non-uniform rates and proposes balancing signal-to-noise ratio across frequencies for more uniform treatment during diffusion.

Further work has moved away from the white noise assumption by exploring structured data corruption methods to preserve image structure. One direction involves heat dissipation~\cite{DBLP:conf/iclr/RissanenHS23}, where the model is trained by stochastically reversing the heat equation to explicitly incorporate the multi-scale inductive biases of natural images. Another direction~\cite{DBLP:conf/nips/BansalBCLKHGGG23} uses completely deterministic destruction of images (e.g., blur, noise, mask and snow). However, these approaches typically lack the theoretical grounding to leverage the natural structure inherent in data. In contrast, PreSS maintains a principled connection between the noise distribution and the power-law structure of spectral components, offering both theoretical justification and empirical benefits.

\section{Preliminaries}

\subsection{Denoising Diffusion Probabilistic Models}
DDPMs consist of two main components: a forward (diffusing) process and a backward (denoising) process. The forward process \cite{ho2020denoising} is defined by a Markov chain parameterized with Gaussian transitions that gradually add Gaussian noise $\mathbf{{\epsilon}_t}$ to progressively corrupt any image data $\mathbf{x}_0\sim p_{data}(\mathbf{x})$, creating increasingly noisy data points. For any $t\in[0,T]$, the latent variable $\mathbf{x}_t$ is defined as:
\begin{equation}\label{eq:DDPM-forward}
    \mathbf{x}_t = \sqrt{\bar{\alpha}_t}\mathbf{x}_0 + \sqrt{1-\bar{\alpha}_t}\mathbf{{\epsilon}}_t,
\end{equation}   
where $\mathbf{{\epsilon}}_t\sim \mathcal{N}(\mathbf{0},\ \mathbf{I})$ and $\mathbf{\bar{\alpha}}_t:=\prod_{s = 1}^{t} {\alpha}_s$. The denoising framework is then trained to invert this process by applying a reverse Markov chain starting from $\mathbf{x}_T \sim \mathcal{N}(\mathbf{0},\ \mathbf{I})$ with a noise predictor $\mathbf{{\epsilon}}_\theta(\mathbf{x}_t,t)$:
\begin{equation}
    \mathbf{x}_{t-1} = \frac{1}{\sqrt{1-\beta_t}}\left[\mathbf{x}_t -\frac{\beta_t}{\sqrt{1-\bar{\alpha}_t}}\mathbf{{\epsilon}}_\theta(\mathbf{x}_t,t)\right]+\sqrt{\beta_t}\mathbf{z},
\end{equation}
where $\mathbf{z}\sim \mathcal{N}(\mathbf{0},\ \mathbf{I})$ and $\mathbf{\beta}_t := 1-{\alpha}_t$. As noted in~\cite{ho2020denoising}, using Bayes theorem, we calculate the posterior distribution of $\mathbf{x}_{t-1}$, denoted by $ q(\mathbf{x}_{t-1}|\mathbf{x}_t,\mathbf{x}_0)$ as defined below: 
\begin{equation}
q(\mathbf{x}_{t-1}|\mathbf{x}_t,\mathbf{x}_0)=\mathcal{N}(\tilde{\boldsymbol{\mu}}_t, \tilde{\beta}_t\mathbf{I}),
\end{equation}
where
\begin{equation}
\tilde{\boldsymbol{\mu}}_t:=\frac{\sqrt{\alpha_t}(1-\bar{\alpha}_{t-1})}{1-\bar{\alpha}_t} \mathbf{x}_t+\frac{\sqrt{\bar{\alpha}_{t-1}}\beta_t}{1-\bar{\alpha}_t}\mathbf{x}_0 ,
\end{equation}
\begin{equation}\label{eq:beta_tilde}
\tilde{\beta}_t:=\frac{1-\bar{\alpha}_{t-1}}{1-\bar{\alpha}_t}\beta_t.
\end{equation}

\subsection{Fourier Transform}
 Fourier transform projects a data point $\mathbf{x}_t(\mathbf{s})\in C(\mathbb{R}^2)$ from the continuous function (pixel) space into the spectral (Fourier) space. The transformation decomposes the image into a weighted sum of trigonometric basis functions, producing spectral (Fourier) components containing the amplitude and phase of each spatial frequency. Formally, the Fourier transform $\mathscr{F}$ is defined by an operator that maps input data $\mathbf{x}_t(\mathbf{s})$ to its corresponding \textit{K}-dimensional spectral components, as shown below: 

\begin{equation}\label{eq:Fourier_Transform}
    \begin{aligned}
        \mathscr{F}:C(\mathbb{R}^2)&\to\mathbb{C}^K \\
        \mathbf{x}_t(\mathbf{s})&\mapsto  \hat{\mathbf{x}}_t,
    \end{aligned}
\end{equation}

where $\mathscr{F}$ denotes the Fourier transform operator, and $\hat{\mathbf{x}}_t=(\hat{x}_{t,1},\cdots,\hat{x}_{t,K})^T$ are multivariate spectral components. More specifically, the spectral composition, i.e. the Fourier transform  $\mathscr{F}$ that connects the spatial variable $\mathbf{s}\in\mathbb{R}^2$ with the wavenumber $\mathbf{k}_j\in\mathbb{R}^2$ is shown through the following integral: 

\begin{equation}
\hat{x}_{t,j} = \mathscr{F}[\mathbf{x}_t(\mathbf{s})] 
= \int_{\mathbb{R}^2} \mathbf{x}_t(\mathbf{s}) \, e^{-i \mathbf{k}_j^T \mathbf{s}} \, d\mathbf{s},
\label{eq:fourier_transform}
\end{equation}

where $e^{-i\mathbf{k}_j^T \mathbf{s}}=\cos(\mathbf{k}_j^T \mathbf{s})-i\sin(\mathbf{k}_j^T \mathbf{s})$, $i$ is the imaginary unit and 
$\mathbf{k}_j$ represents the frequency of the trigonometric functions. Spectral components contain structural information of low-frequency components (global, large-scale structure) and high-frequency components (local, small-scale details). This spectral representation provides a fully interpretable and mathematically tractable space in which statistical information can be calculated and preserved during the diffusion process. 

The spectral decomposition of the input data $\mathbf{x}_t(\mathbf{s})\in C(\mathbb{R}^2)$ is then defined by a linear combination of spatial basis functions and spectral components~\cite{wikle1999dimension, sigrist2015stochastic}: 

\begin{equation}\label{eq:fourier_transform_inv}
  \mathbf{x}_t(\mathbf{s})\approx \sum_{j=1}^K \hat{x}_{t,j} f_j(\mathbf{s}) \equiv \boldsymbol{f}^T(\mathbf{s})\hat{\mathbf{x}}_t,
\end{equation}

where $f_j(\mathbf{s})=\exp(i\mathbf{k}_j^T \mathbf{s})$ is the corresponding spatial basis function, with $\mathbf{s}\in\mathbb{R}^2$ representing the spatial variable and $\mathbf{k}_j$ representing the spatial wavenumbers (frequency vectors). Collectively, these form the Fourier basis, denoted by 
$\boldsymbol{f}(\mathbf{s})=(f_1(\mathbf{s}),\cdots,f_K(\mathbf{s}))^T$, which spans the entire pixel space $C(\mathbb{R}^2)$. This bi-directional mapping using Fourier transform ensures that every dataset within the pixel space can be equivalently represented in an interpretable spectral space.

\section{Methodology}
\begin{table}[h]
\centering
\caption{Notation used throughout the paper.}
\label{tab:notation}
\small
\setlength{\tabcolsep}{6pt}
\renewcommand{\arraystretch}{1.15}
\begin{tabular}{ll}
\toprule
\textbf{Symbol} & \textbf{Description} \\
\midrule
$\mathbf{x}$            & Pixel-space image data \\
$\hat{\mathbf{x}}$      & Spectral components of $\mathbf{x}$ \\
$\mathbf{x}_t$           & Noisy image at diffusion step $t$ \\
$\hat{\mathbf{x}}_t$     & Noisy spectral components at step $t$ \\
$\hat{\boldsymbol{\mu}}$ & Mean values of $\hat{\mathbf{x}}$ \\
$\hat{\boldsymbol{\Sigma}}$ & Variance of $\hat{\mathbf{x}}$ \\
\bottomrule
\end{tabular}
\label{tab:notation_main}
\end{table}

\begin{figure*}[!t]
    \begin{center}
        \includegraphics[width=0.7\linewidth]{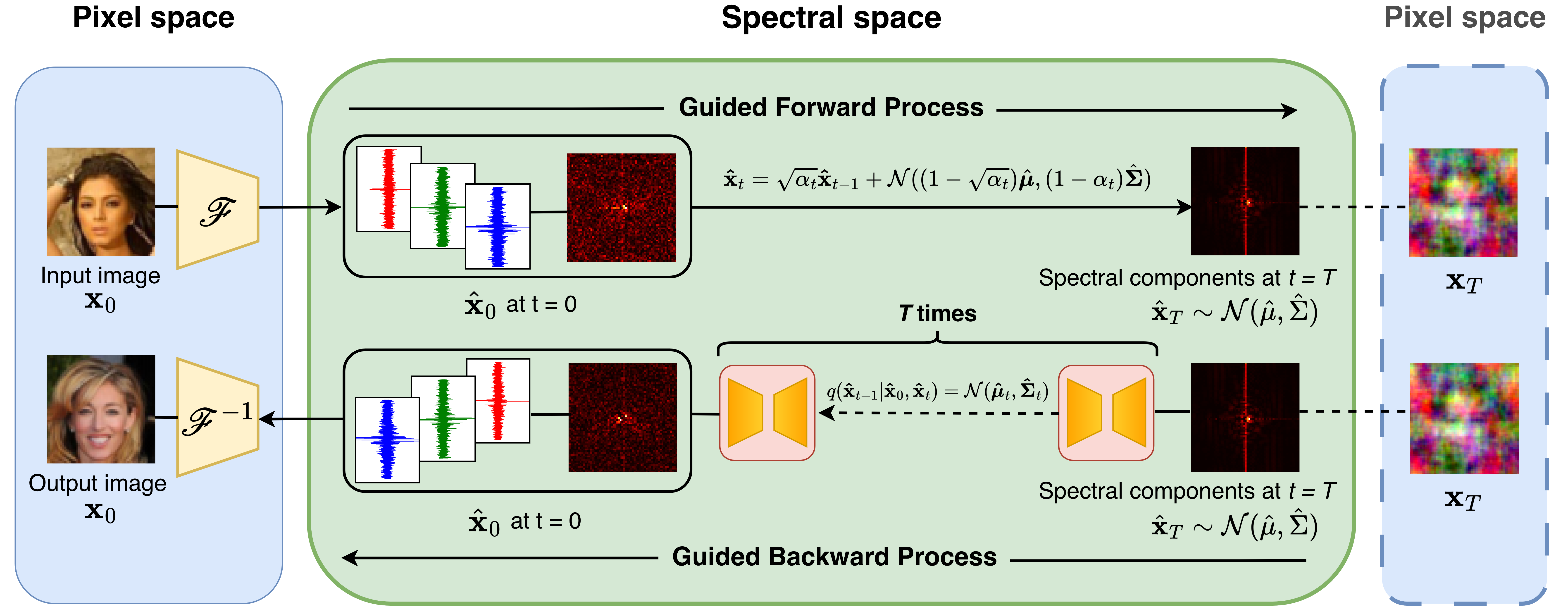}
    \end{center}
    \caption{Overview of our \textbf{Pre}serving Spectral \textbf{S}tructure and \textbf{S}tatistics (\textbf{PreSS}) framework. The input image is transformed into the spectral space via a Fourier transform. Unlike conventional diffusion models which targets a white Gaussian noise, our guided forward process evolves the spectral components towards a statistical prior. The guided backward process then recovers the structural features in the spectral space before mapping back to the pixel space via an inverse Fourier transform.}
    \label{fig:overview}
\end{figure*}

In this section, we reformulate the standard diffusion framework by operating in the spectral space and
guiding both the forward and backward processes. The core motivation of our formulation is to preserve the spectral features of image data and use statistical priors to guide the generative process.  

\subsection{Fourier Transform into Spectral Space}


As we look to preserve spectral features that correspond to power law distribution, Fourier transform is used to project pixel space data into spectral space where these statistical features are observable. For any initial data $x_0$, we obtain its spectral components as follows: 
\begin{equation}
    \mathbf{\hat{x}}_{0} = \mathscr{F}\mathbf{x}_0,
\end{equation}
where $\mathscr{F}:C(\mathbb{R}^2)\to\mathbb{C}^K$ denotes the Fourier transform defined in Eq.~(\ref{eq:Fourier_Transform}).

\subsection{Calculating an Informative Prior} 

Empirical analysis reveals that the distribution of spectral components of natural images is well approximated by a Gaussian distribution. Thus, we can define their structural representation by their mean $\hat{\boldsymbol{\mu}}$ and variance $\hat{\boldsymbol{\Sigma}}$. Given this property, we define an informative spectral prior by calculating the mean $\hat{\boldsymbol{\mu}}$ and variance $\hat{\boldsymbol{\Sigma}}$ across spectral components $\hat{\mathbf{x}}_0$ of all input images $\mathbf{x}_0$. 

This informative spectral prior serves as a foundation to guide our forward process towards and our backward process from a specific random Gaussian distribution $\mathcal{N}(\hat{\boldsymbol{\mu}},\hat{\boldsymbol{\Sigma}})$ as described in Proposition~\ref{prop:forward} and \ref{prop:backward_process}, which ensures that spectral features are preserved throughout the diffusion process.

\subsection{Forward Process in PreSS Diffusion Model}

Formally, we denote the mean of $\mathbf{\hat{x}}_0$ as $\hat{\boldsymbol{\mu}} = (\hat{\mu}_1, \cdots,\hat{\mu}_K)^T$ and the variance of $\mathbf{\hat{x}}_0$ as $\hat{\boldsymbol{\Sigma}}=diag\{\hat{\sigma}_1^2,\cdots,\hat{\sigma}_K^2\}$. We then define the single-step forward process by the proposition below:
\begin{proposition}\label{prop:forward}
Suppose $\mathbf{x}_0$ is sampled from the data distribution $p_{data}(x)$ and $\mathbf{\hat{x}}_0$ are its spectral components, then $\mathbf{\hat{x}}_t$ ($t>0)$ is defined by
\begin{equation}\label{eq:forward}
    \mathbf{\hat{x}}_t = \sqrt{\alpha_t} \mathbf{\hat{x}}_{t-1} + \mathcal{N}((1-\sqrt{\alpha_t})\hat{\boldsymbol{\mu}}, (1-\alpha_t)\hat{\boldsymbol{\Sigma}}).
\end{equation}
\end{proposition}

This formulation implies a closed-form (one-step) expression of forward process defined by:
\begin{equation}\label{eq:closed_forward}
    q(\mathbf{\hat{x}}_t|\mathbf{\hat{x}}_0)=\mathcal{N}(\sqrt{\bar{\alpha}_{t}}\mathbf{\hat{x}}_0
    + (1-\sqrt{\bar{\alpha}_t})\hat{\boldsymbol{\mu}},\ (1-\bar{\alpha}_t)\hat{\boldsymbol{\Sigma}}).
\end{equation}
The closed-form expression in Eq. (\ref{eq:closed_forward}) allows us to sample the spectral components $\mathbf{\hat{x}}_t$ at any arbitrary timestep $t$ directly from initial data $\mathbf{\hat{x}}_0$. 


We then find that at the final timestep $t=T$ when the noise-scheduler satisfies $\alpha_t=\bar{\alpha}_t=0$, the distribution of the closed-form forward process $q(\mathbf{\hat{x}}_T|\mathbf{\hat{x}}_0)$ simplifies to:

\begin{equation}
    q(\mathbf{\hat{x}}_T|\mathbf{\hat{x}}_0)=\mathcal{N}(\hat{\boldsymbol{\mu}},\ \hat{\boldsymbol{\Sigma}}).
\end{equation}

Thus, at the end of the forward process, the input data $\mathbf{\hat{x}}_0$ is diffused into a random noise $\mathcal{N}(\hat{\boldsymbol{\mu}},\ \hat{\boldsymbol{\Sigma}})$ that satisfies a Gaussian distribution. This ensures that the resulting Gaussian noise at terminal time maintains the same mean and variance as the initial spectral components, hence successfully preserving spectral features by preserving spectral mean and variance.

\subsection{Backward Process in PreSS Diffusion Model}

The Gaussian distribution property of data within the spectral space is further established by the premise that the forward process forms a Gaussian Markov chain, which, combined with the additive property of Gaussian distribution ensures that intermediate noisy data $\mathbf{\hat{x}}_t$ remain Gaussian. 

Hence, the posterior distribution $q(\mathbf{\hat{x}}_{t-1}|\mathbf{\hat{x}}_0,\mathbf{\hat{x}}_t)$ necessarily retains Gaussianity in spectral space. We therefore define the backward process by the posterior distribution of $\hat{\mathbf{x}}_{t-1}$:

\begin{proposition}\label{prop:backward_process}
    Given the noisy data $\mathbf{\hat{x}}_t$ and initial data $\mathbf{\hat{x}}_0$, $\hat{\mathbf{x}}_{t-1}$ satisfy the following posterior distribution:
    \begin{equation}
    \label{eq:backward_posterior}
    q(\mathbf{\hat{x}}_{t-1}|\mathbf{\hat{x}}_0,\mathbf{\hat{x}}_t) = \mathcal{N}(\boldsymbol{\hat{\mu}}_t, \boldsymbol{\hat{\Sigma}}_t),
\end{equation}
with mean
\begin{equation}
    \label{eq:backward_posterior_mean}
    \boldsymbol{\hat{\mu}}_t= \frac{\sqrt{\alpha_t}(1-\bar{\alpha}_{t-1})}{1-\bar{\alpha}_t} \mathbf{\hat{x}}_t+\frac{\sqrt{\bar{\alpha}_{t-1}}\beta_t}{1-\bar{\alpha}_t}\mathbf{\hat{x}}_0 +\gamma_t\hat{\boldsymbol{\mu}},
\end{equation}
and variance 
\begin{equation}
    \label{eq:backward_posterior_variance}
    \boldsymbol{\hat{\Sigma}}_t = \tilde{\beta}_t \hat{\mathbf{\Sigma}},
\end{equation}
where $\tilde{\beta}_t$ is defined in Eq.~(\ref{eq:beta_tilde}) and the parameter $\gamma_t$ is defined to be:
\begin{equation}
    \gamma_t:=\frac{1-\sqrt{\bar{\alpha}_{t-1}}}{1-\bar{\alpha}_t}\beta_t-\frac{(1-\bar{\alpha}_{t-1})(\sqrt{\alpha_t}-\alpha_t)}{1-\bar{\alpha}_t}.
\end{equation}
\end{proposition}

This simplifies the model as the reverse Gaussian process can be expressed in closed form in terms of $\hat{\boldsymbol{\mu}}$, $\hat{\mathbf{\Sigma}}$ and the noise scheduler. Furthermore, since the mean $\hat{\boldsymbol{\mu}}_t$ is deterministic and only depends on the current state and initial state, we can then define the single-step backward sampling as given by:
\begin{equation}
    \hat{\mathbf{x}}_{t-1}=\hat{\boldsymbol{\mu}}_t + \hat{\mathbf{z}}_t,\quad \hat{\mathbf{z}}_t\sim \mathcal{N}(\mathbf{0}, \mathbf{\hat{\Sigma}_t}),
\end{equation}
where $\hat{\mathbf{z}}_t$ is a sampled random noise. This backward formula enables a consistent noising-denoising process. Derivation of Proposition~\ref{prop:backward_process} can be found in Appendix~\ref{sec:proof_posterior_distribution}.

We design a denoising function by employing a neural network $m_\theta$ operating in pixel space. Given a noisy image $\mathbf{x}_0$, the network predicts the original image as 
\begin{equation}
    \mathbf{x}_0 = m_\theta(\mathbf{x}_t, t).
\end{equation}
At each backward step for any spectral components $\hat{\mathbf{x}}_t$, we first map them back into pixel space by the inverse Fourier transform $\mathbf{x}_{t} = \mathscr{F}^{-1}\mathbf{\hat{x}}_t$, apply the predictor $m_\theta$, and then transform the predicted image $\mathbf{x}_0$ back into spectral space, as follows:
\begin{equation}
\label{eq:denoising_function}
    \mathbf{\hat{x}}_{0} = \mathscr{F} m_\theta(\mathscr{F}^{-1}\mathbf{\hat{x}}_t, t).
\end{equation}

This design leverages the advantages of the well-designed modern neural networks in pixel space. Instead of redesigning a neural network exclusive for spectral space, we apply existing neural networks to maximize our generality and for a fair comparison. Notably, since $\mathbf{\hat{x}}_t$ is Gaussian and $\mathbf{x}_{t} = \mathscr{F}^{-1}\mathbf{\hat{x}}_t$ is a linear combination defined by Eq.~(\ref{eq:fourier_transform_inv}), $\mathbf{x}_{t}$ is also Gaussian, which can reduce the computational complexity of the neural network. The training objective can be defined as:
\begin{equation}
\label{eq:press_loss}
        \mathcal{L}_{PreSS}\;:=\;
\mathbb{E}_{x,\;\epsilon \sim \mathcal{N}(0,1),\; t}
\left[
\left\lVert
\hat{\mathbf{x}}_0 - \hat{\mathbf{x}}_{0,\theta}(\mathbf{\hat{x}}_t, t)
\right\rVert_2^2
\right],
\end{equation}
where $\hat{\mathbf{x}}_{0,\theta}(\mathbf{\hat{x}}_t, t)$ is obtained by Eq.~(\ref{eq:denoising_function}). 

The forward and backward processes are summarized in Alg. \ref{alg:forward} and Alg. \ref{alg:backward}.

\subsection{Fourier transform in discrete pixel space}

Although Fourier transform is defined for functions in continuous space $C(\mathbb{R}^2)$, real-world applications (e.g., in scientific and engineering fields) typically involve real-valued data sampled on a discrete spatial grid, such as images. We assume that an image $\mathbf{x}_t$ in pixel space is defined on an $N_1\times N_2$ regular spatial grid. The rectangular spatial mesh system is defined by a tensor product of two one-dimensional collocation sets:
\begin{equation}
    \mathbb{S}=\mathbf{s}_1 \otimes \mathbf{s}_2,
\end{equation}
where
\begin{equation}
    \begin{aligned}
        \mathbf{s}_1 & =(s_1^{(i)};s_1^{(i)}=\frac{i}{N_1},i=0,\cdots,N_1-1), \\
        \mathbf{s}_2 & =(s_2^{(i)};s_2^{(i)}=\frac{i}{N_2},i=0,\cdots,N_2-1).
    \end{aligned}
\end{equation}

Without any loss of generality, $N_1$ and $N_2$ are assumed to be even numbers. Hence we are able to restrict our attention to two-dimensional discrete Fourier transform, which can be defined as below:
\begin{equation}\label{eq:dfft_complex}
        \hat{\mathbf{x}}_t(k_1,k_2) =\hat{\mathbf{x}}_t^{(R)}(k_1,k_2)-i\hat{\mathbf{x}}_t^{(I)}(k_1,k_2),
\end{equation}
for $k_1=-\frac{N_1}{2}+1,-\frac{N_1}{2}+2,\cdots,\frac{N_1}{2}$ \\
\smallskip
and $k_2=-\frac{N_2}{2}+1,-\frac{N_2}{2}+2,\cdots,\frac{N_2}{2}$. 

The deviation and complete expression of Eq.~(\ref{eq:dfft_complex}) are provided in Appendix~\ref{sec:Deviation_DFT}. 

Since image $\mathbf{x}_t$ is real-valued, we are able to determine the following from rotational symmetry:
\begin{equation}
    \begin{aligned}
        |\hat{\mathbf{x}}_t(k_1,k_2)|    & =|\hat{\mathbf{x}}_t(-k_1,-k_2)|,     \\
        \arg \hat{\mathbf{x}}_t(k_1,k_2) & = \arg \hat{\mathbf{x}}_t(-k_1,-k_2).
    \end{aligned}
\end{equation}

where $\arg$ is the argument of complex numbers. Therefore, removing half of the complex plane reduces the dimension of $\hat{\mathbf{x}}_t(k_1,k_2)$ to $N_1\times \frac{N_2}{2}$. By concatenating the real and imaginary parts, we can then convert the spectral components into real-valued components:

\begin{equation}\label{eq:dfft_real}
    \hat{\mathbf{x}}_t(k_1,k_2)=\left\{
    \begin{aligned}
         & \hat{\mathbf{x}}_t^{(R)}(k_1,k_2),\quad k_2\geq 0 \\
         & \hat{\mathbf{x}}_t^{(I)}(k_1,-k_2), \quad k_2<0
    \end{aligned}
    \right.
\end{equation}

which exists in the same dimension as $\mathbf{x}_t\in\mathbb{R}^{N_1\times N_2}$. 




\begin{algorithm}[t]
    \caption{Training algorithm}
    \label{alg:forward}
    \SetKwInOut{Input}{Input}
    \Input{Data samples $S := \{ \mathbf{x}_0^{(i)}\}_{i=1}^N$, number of diffusion steps $T$.}

    $\mathbf{\hat{x}}_{0} \gets \mathscr{F}\mathbf{x}_0$\;

    $\hat{\boldsymbol{\mu}} \gets \mathrm{Mean}(\mathbf{\hat{x}}_{0} )$\;

    $\hat{\boldsymbol{\Sigma}} \gets \mathrm{diag}\!\big\{\mathrm{Var}(\mathbf{\hat{x}}_{0} )\big\}$\;

    \While{not converged}{
    $t\sim \mathrm{Uniform} (\{1,2,\cdots,T\})$
    
    \textbf{Forward process:}

    \ \ $\mathbf{\hat{x}}_t = \sqrt{\alpha_t} \mathbf{\hat{x}}_{t-1} + \mathcal{N}((1-\sqrt{\alpha_t})\hat{\boldsymbol{\mu}}, (1-\alpha_t)\hat{\boldsymbol{\Sigma}})$
    
    \textbf{Predict sample:}

    \ \ $\hat{\mathbf{x}}_{0,\theta} = \mathscr{F} m_\theta(\mathscr{F}^{-1}\mathbf{\hat{x}}_t, t)$
    
    \textbf{Compute loss:}

    \ \ $\mathcal{L}_t = \left\lVert
\hat{\mathbf{x}}_0 - \hat{\mathbf{x}}_{0,\theta}
\right\rVert_2^2$

    Update $\theta$\;
    }
\end{algorithm}

\begin{algorithm}[t]
    \caption{Sampling algorithm}
    \label{alg:backward}
    \SetKwInOut{Input}{Input}
    \Input{Random noise $\mathbf{\hat{x}}_T\sim\mathcal{N}(\hat{\boldsymbol{\mu}},\hat{\boldsymbol{\Sigma}})$}

    \For{$t=T; t\geq 1; t=t-1$}{
        $\mathbf{\hat{x}}_{0} = \mathscr{F} m_\theta(\mathscr{F}^{-1}\mathbf{\hat{x}}_t, t)$ \;

        $\hat{\mathbf{x}}_{t-1}=\hat{\boldsymbol{\mu}}_t + \hat{\mathbf{z}}_t,\quad \hat{\mathbf{z}}_t\sim \mathcal{N}(\mathbf{0}, \mathbf{\hat{\Sigma}_t})$
    }
    \Return $\mathscr{F}^{-1}\mathbf{\hat{x}}_0$
\end{algorithm}

\section{Extension to Other DDPM-based Models}
\label{sec:extension}

By offering a principled alternative to standard Denoising Diffusion Probabilistic Models (DDPMs), our approach is designed for natural integration into the various existing frameworks that are built upon the DDPM backbone. This high degree of architectural compatibility allows PreSS to be incorporated into existing diffusion-based models with minimal modifications, essentially serving as a "plug-and-play" enhancement for various generative pipelines. By doing so, PreSS not only elevates the performance of current state-of-the-art models but also establishes a more robust foundation for future innovations in reducing computational complexity and high-resolution image generation. 

In this section, we demonstrate how PreSS can be naturally extended to other DDPM-based models.

\subsection{Extension to Latent Diffusion Models}

\begin{figure}[h]
    \begin{center}
        \includegraphics[width=1.0\linewidth]{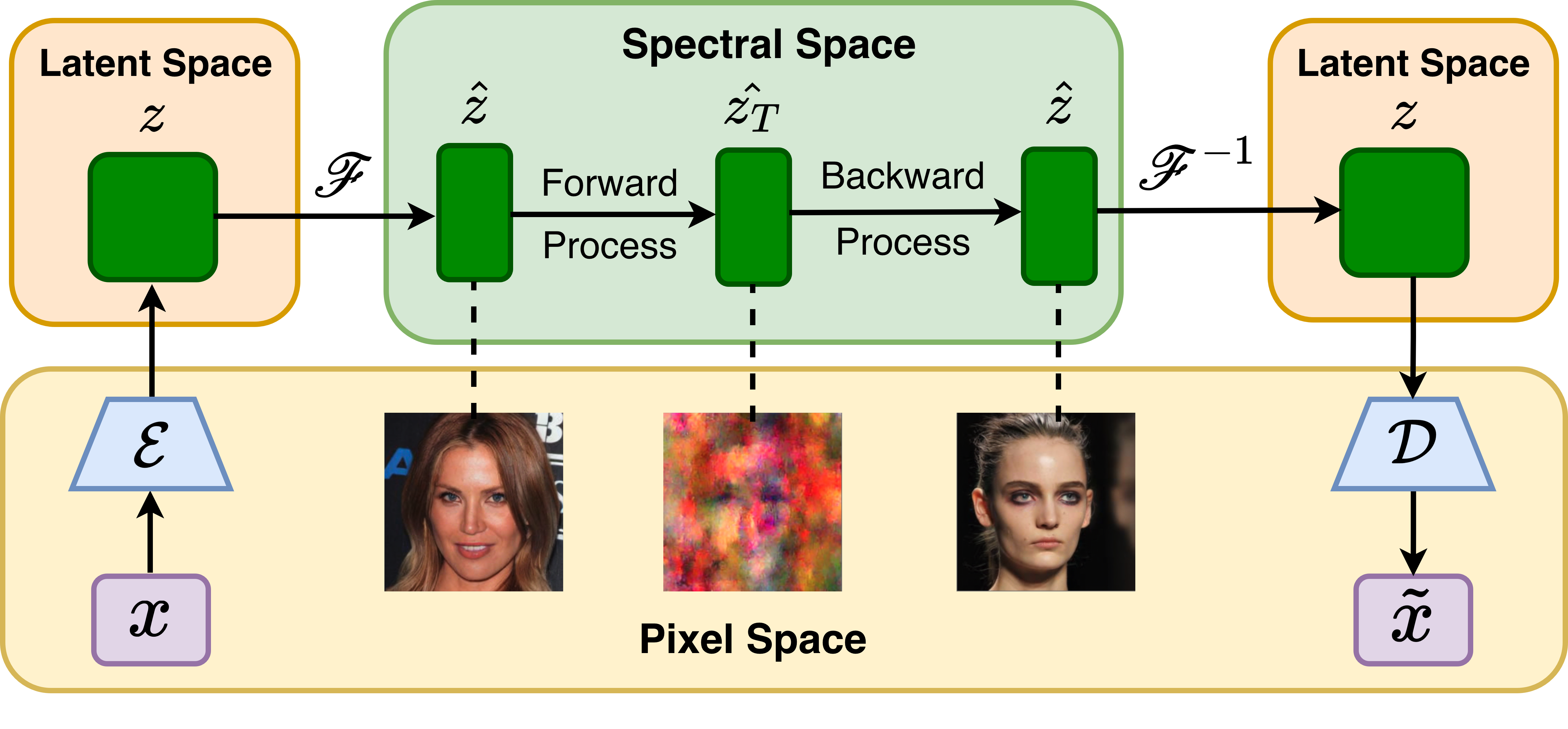}
    \end{center}
    \caption{\textbf{Illustration of PreSS applied to LDM.} Diffusion process is performed within the spectral space of latent variables to preserve spectral statistics across the forward and backward processes, while maintaining smooth image evolution after decoding to pixel space.}
    \label{fig:press_ldm}
\end{figure}

\textbf{Latent Diffusion Models (LDMs).} As a representative example, we demonstrate the extensibility of PreSS by applying it to {Latent Diffusion Models (LDMs)}~\cite{DBLP:conf/cvpr/RombachBLEO22}, a widely adopted framework for high-resolution image synthesis. LDM operates by first encoding an input data $\mathbf{x}_0$ into a latent representation $\mathbf{z}_0$ via a pre-trained autoencoder $\mathcal{E}$. The diffusion process is then trained directly on this latent representation (rather than a pixel space) to denoise $\mathbf{z}_t$ instead of $\mathbf{x}_t$. The training objective of LDM is given by:
\begin{equation}
    \mathcal{L}_{\mathrm{LDM}}\;:=\;
\mathbb{E}_{\mathcal{E}(x),\;\epsilon \sim \mathcal{N}(0,1),\; t}
\left[
\left\lVert
\epsilon - \epsilon_\theta(\mathbf{z}_t, t)
\right\rVert_2^2
\right],
\end{equation}

where $\mathbf{z}_t$ is the noisy latent at an arbitrary timestep $t$, derived by adding Gaussian noise to $\mathbf{z}_0$ according to the standard diffusion model framework as shown in Eq.~(\ref{eq:DDPM-forward}).

\textbf{Extending PreSS to LDM.} We operate on the latent variables within spectral space. Specifically, we apply Fourier transform to the latent variable $\mathbf{z}_0$ to obtain the spectral components $\hat{\mathbf{z}}_0$ and calculate their corresponding mean and variance in the spectral space. The forward and backward diffusion processes are then carried out in the spectral space as defined in Proposition~\ref{prop:forward} and \ref{prop:backward_process}. The denoised spectral components $\hat{\mathbf{z}}_0$ are then transformed back to the latent variable $\mathbf{z}_0$ and subsequently decoded into pixel space by the pre-trained decoder $\mathcal{D}$. The resulting training objective is formulated as:
\begin{equation}
        \mathcal{L}_{PreSS}\;:=\;
\mathbb{E}_{\mathcal{E}(x),\;\epsilon \sim \mathcal{N}(0,1),\; t}
\left[
\left\lVert
\hat{\mathbf{z}}_0 - \hat{\mathbf{z}}_{0,\theta}(\hat{\mathbf{z}}_t, t)
\right\rVert_2^2
\right].
\end{equation}

\begin{figure}[h]
    \begin{center}
        \includegraphics[width=0.8\linewidth]{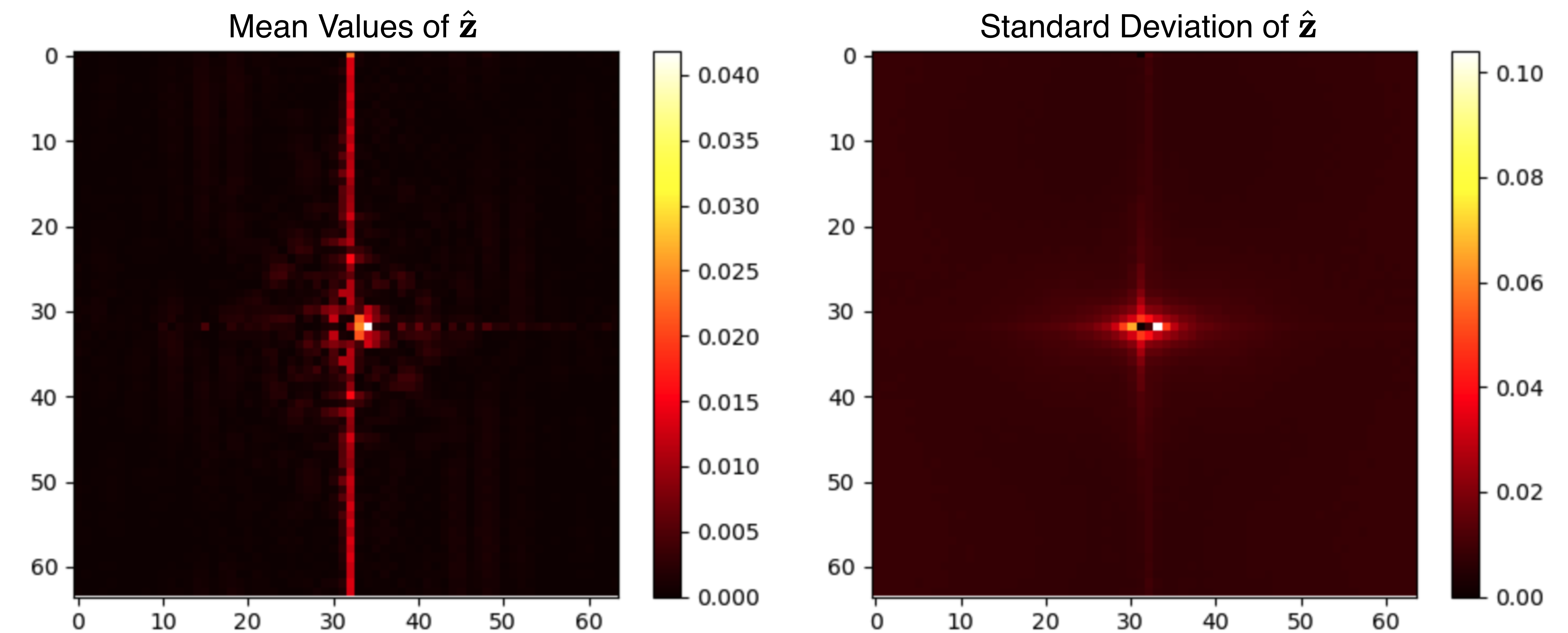}
    \end{center}
    \caption{Heatmap of the mean (left) and standard deviation (right) of $\hat{\mathbf{z}}$, where $\mathbf{z}$ is obstained by a pre-trained VQ-VAE encoder.}
    \label{fig:press_ldm_mean_std}
\end{figure}

This LDM-PreSS integration framework is further illustrated in Figure~\ref{fig:press_ldm}. Notably, Figure~\ref{fig:press_ldm_mean_std} reveals that the encoded variables $\hat{\mathbf{z}}_0$ contains spectral structure and statistical information (specifically mean and variance). By operating within a spectral space, PreSS is capable of leveraging statistical priors unexplored by conventional diffusion models. As a result, this extension enables the enhancement of visual diversity and image generation quality by PreSS framework, while also applying the advantages of the LDM framework. 


The successful implementation of PreSS within the LDM framework demonstrates our model's robust integration to DDPM-based models without necessitating structural redesigns. Ultimately, this demonstrates that our approach is not a standalone modification, but a generalizable improvement on the diffusion objective that scales across DDPM-based models. 

\subsection{Extension to Conditional Generation}
The PreSS framework can also be readily extended to conditional generation tasks, where the objective is to generate data samples conditioned on specific attributes or inputs $\mathbf{c}$. In this setting, 
the prior distribution of the spectral components is calculated based on the conditional data distribution. 

In the training process, each input data $\mathbf{x}_0$ is paired with its corresponding condition $\mathbf{c}$. We calculate the conditional mean and variance of $\hat{\mathbf{x}}_0$ as:
\begin{equation}
    \hat{\boldsymbol{\mu}}_{\mathbf{c}}:=\mathbb{E}\left[\hat{\mathbf{x}}_0 | \mathbf{c}\right],\quad \hat{\boldsymbol{\Sigma}}_{\mathbf{c}}:=\text{Var}\left[\hat{\mathbf{x}}_0 | \mathbf{c}\right].
\end{equation}

The forward process progressively destroyed $\hat{\mathbf{x}}_0$ into its corresponding Gaussian noise $\mathcal{N}(\hat{\boldsymbol{\mu}}_c,\ \hat{\boldsymbol{\Sigma}}_c)$. The diffusion model is then trained to denoise $\hat{\mathbf{x}}_t$ while preserving the spectral structure and statistics conditioned on $\mathbf{c}$. The training objective can be formulated as:
\begin{equation}
    \mathcal{L}_{\text{cond}} = \mathbb{E}_{\mathbf{x}_0,\mathbf{c},\epsilon,t}
\left[
\left\lVert
\hat{\mathbf{x}}_0 - \hat{\mathbf{x}}_{0,\theta}(\hat{\mathbf{x}}_t, \mathbf{c}, t)
\right\rVert_2^2
\right].
\end{equation}

Finally, during the sampling process, the model generates samples by iteratively denoising $\hat{\mathbf{x}}_t$ while conditioning on $\mathbf{c}$, ensuring that the generated samples satisfy the desired attributes while maintaining their spectral structure. This extension enables PreSS to be effectively applied to a wide range of conditional generation tasks, including class-conditional image generation and text-to-image generation.

\section{Experiments}
\label{sec:experiments}

Given the theoretical formulations of PreSS, we then look to evaluate its effectiveness in practice. In particular, we look to evaluate whether our proposed diffusion process in spectral space preserves spectral structure and statistics, and whether such preservation leads to improved image generation performance. 


\paragraph{Experiment Settings}
Our experiments are conducted on three widely adopted datasets with different resolutions: CIFAR-10 ($32\times32$)~\cite{krizhevsky2009learning}, CelebA ($64\times64$)~\cite{liu2015faceattributes}, and CelebA-HQ ($256\times 256$)~\cite{karras2017progressive}. Furthermore, we adopted two standard metrics, Fréchet Inception Distance (FID) and Inception Score (IS), to quantitatively evaluate the generation quality and diversity. As baselines, we compare PreSS with two representative diffusion frameworks: Denoising Diffusion Probabilistic Models (DDPM)~\cite{ho2020denoising} using a cosine noise schedule~\cite{DBLP:conf/icml/NicholD21} and Latent Diffusion Models~\cite{DBLP:conf/cvpr/RombachBLEO22} for experiments on CelebA-HQ.

\subsection{Performance on Unconditional Generation}

\begin{table}[h]
\centering
\caption{Quantitative comparison between DDPM, LDM and PreSS.
Arrows indicate whether higher or lower values are better.}
\small
\setlength{\tabcolsep}{3.5pt}   
\begin{tabular}{lcc|cc|cc}
\toprule
\multirow{2}{*}{Metric}
& \multicolumn{2}{c|}{CIFAR-10}
& \multicolumn{2}{c|}{CelebA}
& \multicolumn{2}{c}{CelebA-HQ} \\
& DDPM & PreSS
& DDPM & PreSS
& LDM & PreSS \\
\midrule
FID ($\downarrow$)     & 3.05 & \textbf{2.89} & 4.58 & \textbf{3.97} & 5.11 & \textbf{4.62} \\
IS ($\uparrow$)        & 9.07 & \textbf{9.70}  & 11.43 & \textbf{12.48}  & 3.29 & \textbf{3.55} \\
\bottomrule
\end{tabular}
\label{tab:quantitative_results_compact}
\end{table}

To evaluate the unconditional generation quality and diversity, a standard U-Net architecture~\cite{DBLP:conf/icml/NicholD21} is employed for all diffusion models to ensure fair comparison. On CIFAR-10 and CelebA, we compare PreSS with improved DDPM~\cite{DBLP:conf/icml/NicholD21} using the $\mathcal{L}_{simple}$ objective. For CelebA-HQ, we use a pre-trained autoencoder and compare PreSS with LDM. 
As shown in Table~\ref{tab:quantitative_results_compact}, \textbf{PreSS consistently outperforms the baseline models across all datasets and metrics}, demonstrating its effectiveness in preserving spectral features while improving generation quality and diversity. Notably, the performance gains on CelebA are more pronounced than the performance on CIFAR-10, likely due to the richer structural and statistical information in higher-resolution facial images. 

\begin{figure*}[t]
    \begin{center}
        \includegraphics[width=1.0\linewidth]{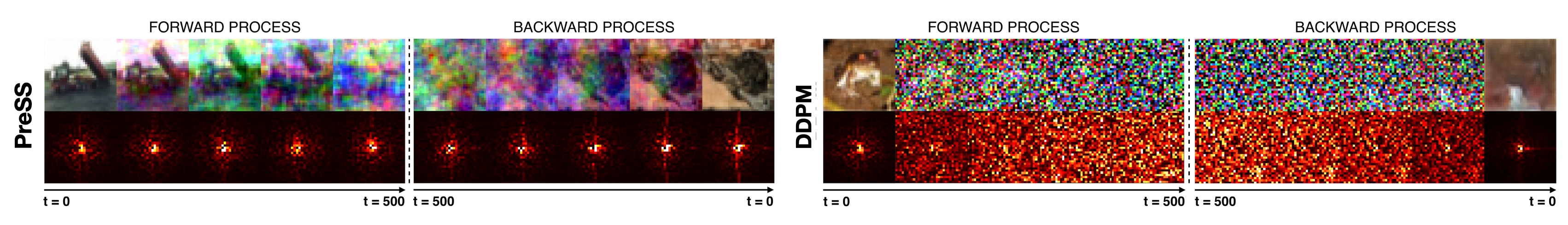}
    \end{center}
    \caption{Visual comparison of forward and backward processes between PreSS (\textbf{left}) and DDPM (\textbf{right}) by illustration of the images (\textbf{top row}) with their corresponding spectral components (\textbf{bottom row}) at different diffusion time-steps. } 
    \label{fig:cifar10_stepbystep}
\end{figure*}



\subsection{Ablation Study}

To validate the contribution of our proposed model, we conduct an ablation study by comparing PreSS with a variant model that applies the standard DDPM directly into spectral space via Fourier transform. This variant model diffuses spectral components toward white noise and destroys spectral features. As shown in Table~\ref{tab:ablation_results}, PreSS significantly outperforms the variant model across all metrics on CIFAR-10. While the variant model achieves similar FID and IS to DDPM, it fails to benefit from spectral space modeling. 
These results demonstrate that explicitly preserving spectral structure and statistics during the diffusion process is crucial for achieving superior generation quality and diversity.

\begin{table}[h]
\centering
\caption{Quantitative comparison between DDPM, DDPM(with Fourier transform) and PreSS on CIFAR-10 dataset. Arrows indicate whether higher or lower values are better.}
\small
\begin{tabular}{lc|c|c}
\toprule
Metric
& DDPM & DDPM(FT) & PreSS \\
\midrule
FID ($\downarrow$)     & \underline{3.05} & 3.09 & \textbf{2.89}  \\
IS ($\uparrow$)        & 9.07 & \underline{9.12} & \textbf{9.70} \\
\bottomrule
\end{tabular}
\label{tab:ablation_results}
\end{table}




\subsection{Analysis of Frequency Invariance in Spectral Space}

To further analyze whether PreSS can effectively preserve the spectral features, we visualize the forward and backward processes of PreSS and DDPM in both pixel space and spectral space for diffusion steps ranging from $t=0$ to $t=500$. As illustrated in Figure~\ref{fig:cifar10_stepbystep}, PreSS maintains stable spectral structure throughout the diffusion process, with the distributions of spectral components remaining largely unchanged and evolving smoothly over time. In contrast, DDPM rapidly diffuses the image and destroys spectral structure, resulting in noisy spectral components and severe corruption of statistical information. By retaining spectral structure, PreSS enables a more stable and smooth diffusion process and facilitates the reconstruction of high-frequency spectral components corresponding to fine image details. This stability leads to improved generation quality while maintaining diversity, highlighting the benefits of modeling the diffusion process in the spectral space and preserving spectral features.

\subsection{Performance on Conditional Generation}

\begin{figure}[h]
\begin{center}
\includegraphics[width=0.8\linewidth]{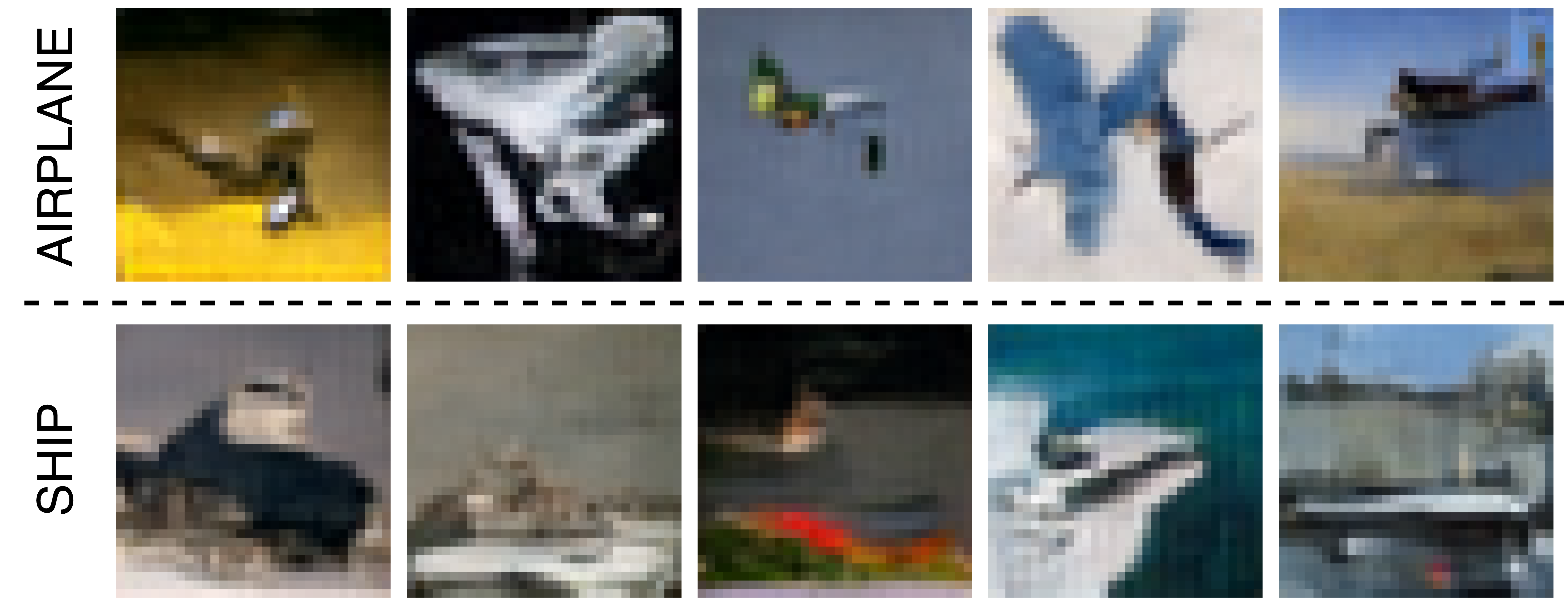}
\end{center}
\caption{Class-conditional generation results on CIFAR-10. We specify two target classes (airplane and ship) and generate samples starting from class-related random noises. The generated images are well-aligned with the specified classes, demonstrating effective controllable class-conditional generation. 
}
\label{fig:results_cifar10}
\end{figure}

To evaluate the controllability of PreSS, we perform class-conditional image generation on the CIFAR-10 dataset with spectral mean $\hat{\boldsymbol{\mu}}_\mathbf{c}$ and variance $\hat{\boldsymbol{\Sigma}}_\mathbf{c}$ conditional on class. During the training process, the denoising function $m_\theta$ itself remains unconditional. During sampling, PreSS starts from random Gaussian noise $\mathcal{N}(\hat{\boldsymbol{\mu}}_\mathbf{c},\ \hat{\boldsymbol{\Sigma}}_\mathbf{c})$ corresponding to the target class. Figure~\ref{fig:results_cifar10} shows representative samples generated for two classes (airplane and ship). PreSS generates images consistent with the specified categories, demonstrating its effective class-conditional controllability. Despite starting from a random noise, the model reliably generates samples with correct classes, suggesting that class-specific spectral statistics contain meaningful conditional information. This observation highlights the potential of leveraging spectral structure and statistics for stable and controllable generation. Additional conditional generation results are provided in Appendix~\ref{sec:conditional_generation_results}.

\newcommand{\best}[1]{\textbf{#1}} 
\section{Conclusion}


In this work, we discovered and analyzed the structure and statistical information in the spectral space. Specifically, we empirically verified the existence of spectral structure and statistics. Building on this insight, we proposed PreSS, a diffusion model that preserves the mean and variance of spectral components during the forward and backward processes. Extensive experiments on CIFAR-10, CelebA and CelebA-HQ datasets demonstrated that PreSS not only maintains the spectral structure and statistics, but also effectively improves the generation quality and diversity. The supplementary materials provide discussions on additional experimental results and our limitations.

\bibliography{main}
\bibliographystyle{icml2026}

\newpage
\appendix
\onecolumn
\clearpage

\section{Theoretical Proofs and Discussions}

\subsection{Deviation of Backward Posterior Distribution}
\label{sec:proof_posterior_distribution}
In this section, we will show the derivation of Eq.~(\ref{eq:backward_posterior}). Recall that the forward process is:

\begin{equation}
    \mathbf{\hat{x}}_t = \sqrt{\alpha_t} \mathbf{\hat{x}}_{t-1} + \mathcal{N}((1-\sqrt{\alpha_t})\hat{\boldsymbol{\mu}}, (1-\alpha_t)\hat{\boldsymbol{\Sigma}})
\end{equation}
Given $\mathbf{\hat{x}}_0$, the distribution of $\mathbf{\hat{x}}_t$ is:
\begin{equation}
    q(\mathbf{\hat{x}}_t|\mathbf{\hat{x}}_0)=\mathcal{N}(\sqrt{\bar{\alpha}_{t}}\mathbf{\hat{x}}_0
    + (1-\sqrt{\bar{\alpha}_t})\hat{\boldsymbol{\mu}},\ (1-\bar{\alpha}_t)\hat{\boldsymbol{\Sigma}})
\end{equation}
By Bayes' theorem, the posterior distribution can be written as:
\begin{equation}\label{eq:posterior-prop}
    q(\mathbf{\hat{x}}_{t-1}|\mathbf{\hat{x}}_0, \mathbf{\hat{x}}_{t}) \propto q(\mathbf{\hat{x}}_t|\mathbf{\hat{x}}_{t-1})q(\mathbf{\hat{x}}_{t-1}|\mathbf{\hat{x}}_0)
\end{equation}
Note that $q(\mathbf{\hat{x}}_t|\mathbf{\hat{x}}_{t-1})$ and $q(\mathbf{\hat{x}}_{t-1}|\mathbf{\hat{x}}_0)$ are multidimensional Gaussian distributions, which can be expressed as:
\begin{equation}\label{eq:posterior-prop-1}
    \begin{aligned}
        q(\mathbf{\hat{x}}_t|\mathbf{\hat{x}}_{t-1}) & \propto \exp\left\{-\frac{1}{2(1-\alpha_t)\hat{\boldsymbol{\Sigma}}}\|\mathbf{\hat{x}}_t-\sqrt{\alpha_t}\mathbf{\hat{x}}_{t-1}-(1-\sqrt{\alpha_t})\hat{\boldsymbol{\mu}}\|^2\right\}                                                                                                          \\
                                                     & \propto \exp\left\{-\frac{1}{2(1-\alpha_t)\hat{\boldsymbol{\Sigma}}}\left[\alpha_t\mathbf{\hat{x}}_{t-1}^T\mathbf{\hat{x}}_{t-1}-2\sqrt{\alpha_t}\mathbf{\hat{x}}_{t-1}^T\mathbf{\hat{x}}_t+2\sqrt{\alpha_t}(1-\sqrt{\alpha_t})\mathbf{\hat{x}}_{t-1}^T\hat{\boldsymbol{\mu}} \right]\right\}
    \end{aligned}
\end{equation}
and
\begin{equation}\label{eq:posterior-prop-2}
    \begin{aligned}
        q(\mathbf{\hat{x}}_{t-1}|\mathbf{\hat{x}}_0) & \propto \exp\left\{-\frac{1}{2(1-\bar{\alpha}_{t-1})\hat{\boldsymbol{\Sigma}}}\|\mathbf{\hat{x}}_{t-1}-\sqrt{\bar{\alpha}_{t-1}}\mathbf{\hat{x}}_0-(1-\sqrt{\bar{\alpha}_{t-1}})\hat{\boldsymbol{\mu}} \|^2 \right\}                                                                                   \\
                                                     & \propto \exp\left\{-\frac{1}{2(1-\bar{\alpha}_{t-1})\hat{\boldsymbol{\Sigma}}}\left[\mathbf{\hat{x}}_{t-1}^T\mathbf{\hat{x}}_{t-1}-2\sqrt{\bar{\alpha}_{t-1}}\mathbf{\hat{x}}_{t-1}^T \mathbf{\hat{x}}_0-2(1-\sqrt{\bar{\alpha}_{t-1}})\mathbf{\hat{x}}_{t-1}^T\hat{\boldsymbol{\mu}}\right] \right\}
    \end{aligned}
\end{equation}
By substituting Eq.~(\ref{eq:posterior-prop-1}) and Eq.~(\ref{eq:posterior-prop-2}) into Eq.~(\ref{eq:posterior-prop}), we have
\begin{equation}
    \begin{aligned}
        q(\mathbf{\hat{x}}_{t-1}|\mathbf{\hat{x}}_0,\mathbf{\hat{x}}_t) & \propto \exp\left\{ -\frac{1}{2\hat{\boldsymbol{\Sigma}}}\left[\mathbf{\hat{x}}_{t-1}^T \left(\frac{\alpha_t}{1-\alpha_t}+\frac{1}{1-\bar{\alpha}_{t-1}}\right)I\mathbf{\hat{x}}_{t-1} \right.\right.                  \\
                                                                        & \qquad\qquad\qquad\quad -2\mathbf{\hat{x}}_{t-1}^T\left((\sqrt{\alpha_t}\mathbf{\hat{x}}_t-\sqrt{\alpha_t}(1-\sqrt{\alpha_t})\hat{\boldsymbol{\mu}})\frac{1}{1-\alpha_t}\right.                                        \\
                                                                        & \qquad\qquad\qquad\quad\quad\quad\quad\quad\left.\left.\left. +(\sqrt{\bar{\alpha}_{t-1}}\mathbf{\hat{x}}_0+(1-\sqrt{\bar{\alpha}_{t-1}})\hat{\boldsymbol{\mu}})\frac{1}{1-\bar{\alpha}_{t-1}}\right) \right] \right\}
    \end{aligned}
\end{equation}
Let $\tilde{\beta}_t^{-1}=\frac{\alpha_t}{1-\alpha_t}+\frac{1}{1-\bar{\alpha}_{t-1}}$, then the posterior distribution can be simplified as:
\begin{equation}
    q(\mathbf{\hat{x}}_{t-1}|\mathbf{\hat{x}}_0,\mathbf{\hat{x}}_t)=\mathcal{N}(\hat{\boldsymbol{\mu}}_t, \hat{\boldsymbol{\Sigma}}_t)
\end{equation}
where
\begin{equation}
    \begin{aligned}
        \hat{\boldsymbol{\Sigma}}_t & = \tilde{\beta}_t\hat{\boldsymbol{\Sigma}},                                                                                                                                                                                                                                                                                                          \\
        \hat{\boldsymbol{\mu}}_t    & =\tilde{\beta}_t\left((\sqrt{\alpha_t}\mathbf{\hat{x}}_t-\sqrt{\alpha_t}(1-\sqrt{\alpha_t})\hat{\boldsymbol{\mu}})\frac{1}{1-\alpha_t}+(\sqrt{\bar{\alpha}_{t-1}}\mathbf{\hat{x}}_0+(1-\sqrt{\bar{\alpha}_{t-1}})\hat{\boldsymbol{\mu}})\frac{1}{1-\bar{\alpha}_{t-1}}\right)                                                                             \\
                                    & = \frac{\sqrt{\alpha_t}(1-\bar{\alpha}_{t-1})}{1-\bar{\alpha}_t}\mathbf{\hat{x}}_t + \frac{\sqrt{\bar{\alpha}_{t-1}}\beta_t}{1-\bar{\alpha}_t}\mathbf{\hat{x}}_0 + \frac{1-\sqrt{\bar{\alpha}_{t-1}}}{1-\bar{\alpha}_t}\beta_t \hat{\boldsymbol{\mu}} - \frac{(1-\bar{\alpha}_{t-1})(\sqrt{\alpha_t}-\alpha_t)}{1-\bar{\alpha}_t}\hat{\boldsymbol{\mu}}.
    \end{aligned}
\end{equation}

\subsection{Asymptotical Discussion of Mean Value}
In this section, we will discuss the asymptotical behavior of the mean value $\hat{\mathbf{x}}_t$ when $t$ is large. Consider a more general case where the forward process is defined as:
\begin{equation}
    \mathbf{\hat{x}}_t = \sqrt{\alpha_t} \mathbf{\hat{x}}_{t-1} + \mathcal{N}(\lambda_t \hat{\boldsymbol{\mu}}, (1-\alpha_t)\hat{\boldsymbol{\Sigma}})
\end{equation}
where $\lambda_t$ is a time-dependent parameter. The closed-form forward process becomes:
\begin{equation}
    q(\mathbf{\hat{x}}_t|\mathbf{\hat{x}}_0)=\mathcal{N}(\sqrt{\bar{\alpha}_{t}}\mathbf{\hat{x}}_0
    + \sqrt{\bar{\alpha}_t}\sum_{s=1}^t\frac{\lambda_t}{\sqrt{\bar{\alpha}_s}} \hat{\boldsymbol{\mu}},\ (1-\bar{\alpha}_t)\hat{\boldsymbol{\Sigma}})
\end{equation}
Note that if $\lambda_t = 1-\sqrt{\alpha_t}$, then it reduces to our original formulation, i.e., Eq~\ref{eq:closed_forward}, whose mean value is:
\begin{equation}
    \mathbb{E}[\mathbf{\hat{x}}_t] = \sqrt{\bar{\alpha}_{t}}\mathbf{\hat{x}}_0
    + (1-\sqrt{\bar{\alpha}_t}) \hat{\boldsymbol{\mu}}
\end{equation}
When $t$ is large enough, $\bar{\alpha}_t \to 0$, thus $\mathbb{E}[\mathbf{\hat{x}}_t] \to \boldsymbol{\mu}$, which means that the mean value is convergent and bounded.
Now we analyze the sufficient condition for the mean value to be convergent and bounded when $t$ is large. Consider $\sqrt{\bar{\alpha}_t}\sum_{s=1}^t\frac{\lambda_t}{\sqrt{\bar{\alpha}_s}}$:
\begin{equation}
    \begin{aligned}
        \frac{\sqrt{\bar{\alpha}_{t+1}}\sum_{s=1}^{t+1}\frac{\lambda_{t+1}}{\sqrt{\bar{\alpha}_s}}}{\sqrt{\bar{\alpha}_t}\sum_{s=1}^t\frac{\lambda_t}{\sqrt{\bar{\alpha}_s}}} & =
        \frac{\sqrt{\bar{\alpha}_{t+1}} \left(\frac{\gamma_1}{\sqrt{\bar{\alpha}_1}}+\cdots+\frac{\gamma_{t+1}}{\sqrt{\bar{\alpha}_{t+1}}}\right)}{\sqrt{\bar{\alpha}_{t}} \left(\frac{\gamma_1}{\sqrt{\bar{\alpha}_1}}+\cdots+\frac{\gamma_{t}}{\sqrt{\bar{\alpha}_{t}}}\right)}                                                                                              \\
                                                                                                                                                                              & =\sqrt{\alpha_{t+1}}+\frac{\sqrt{\alpha_{t+1}}\cdot \frac{\gamma_{t+1}}{\sqrt{\bar{\alpha}_{t+1}}} }{\frac{\gamma_1}{\sqrt{\bar{\alpha}_1}}+\cdots+\frac{\gamma_{t}}{\sqrt{\bar{\alpha}_{t}}}} \\
                                                                                                                                                                              & \leq \sqrt{\alpha_{t+1}}+\frac{\sqrt{\alpha_{t+1}}\cdot \frac{\gamma_{t+1}}{\sqrt{\bar{\alpha}_{t+1}}} }{ \frac{\gamma_{t}}{\sqrt{\bar{\alpha}_{t}}}}                                          \\
                                                                                                                                                                              & = \sqrt{\alpha_{t+1}}+\frac{\gamma_{t+1}}{\gamma_t}
    \end{aligned}
\end{equation}
If there exists a constant $p<1$ such that $\frac{\gamma_{t+1}}{\gamma_t}\leq p<1$, then we have
\begin{equation}
    \begin{aligned}
        \lim_{t\to\infty}\frac{\sqrt{\bar{\alpha}_{t+1}}\sum_{s=1}^{t+1}\frac{\lambda_{t+1}}{\sqrt{\bar{\alpha}_s}}}{\sqrt{\bar{\alpha}_t}\sum_{s=1}^t\frac{\lambda_t}{\sqrt{\bar{\alpha}_s}}} &
        \leq q<1
    \end{aligned}
\end{equation}
By the ratio test, we conclude that $\sqrt{\bar{\alpha}_t}\sum_{s=1}^t\frac{\lambda_t}{\sqrt{\bar{\alpha}_s}}$ is convergent, and thus the mean value $\mathbb{E}[\mathbf{\hat{x}}_t]$ is convergent and bounded when $t$ is large.
In summary, a sufficient condition for the mean value to be convergent and bounded when $t$ is large is that there exists a constant $p<1$ such that $\frac{\lambda_{t+1}}{\lambda_t}\leq p<1$.

\subsection{Deviation of Discrete Fourier Transform Representation}\label{sec:Deviation_DFT}
In this section, we will show the derivation of Eq.~(\ref{eq:dfft_complex}): 
\begin{equation}
    \begin{aligned}
        \hat{\mathbf{x}}_t(k_1,k_2) & =\frac{1}{N_1N_2}\sum_{n_1,n_2} \mathbf{x}_t(n_1,n_2)e^{-i(\frac{2\pi}{N_1}n_1k_1+\frac{2\pi}{N_2}n_2k_2)} \\
                                    & = \frac{1}{N_1N_2}\sum_{n_1,n_2} \mathbf{x}_t(n_1,n_2)e^{-i\omega_{n_1,k_1,n_2,k_2} }                      \\
                                    & =\frac{1}{N_1N_2}\sum_{n_1,n_2} \mathbf{x}_t(n_1,n_2)\cos(\omega_{n_1,k_1,n_2,k_2})-i\frac{1}{N_1N_2}\sum_{n_1,n_2} \mathbf{x}_t(n_1,n_2) \sin(\omega_{n_1,k_1,n_2,k_2})                \\
                                    & =\hat{\mathbf{x}}_t^{(R)}(k_1,k_2)-i\hat{\mathbf{x}}_t^{(I)}(k_1,k_2),
    \end{aligned}
\end{equation}
for $k_1=-\frac{N_1}{2}+1,-\frac{N_1}{2}+2,\cdots,\frac{N_1}{2}$, $k_2=-\frac{N_2}{2}+1,-\frac{N_2}{2}+2,\cdots,\frac{N_2}{2}$, $n_1=0,\cdots,N_1-1$ and $n_2=0,\cdots,N_2-1$.

\section{Experimental Results}

\subsection{Performance on Conditional Generation}
\label{sec:conditional_generation_results}
\begin{figure*}[t]
    \centering
    \begin{subfigure}[b]{0.33\textwidth}
        \centering
        \includegraphics[width=\linewidth]{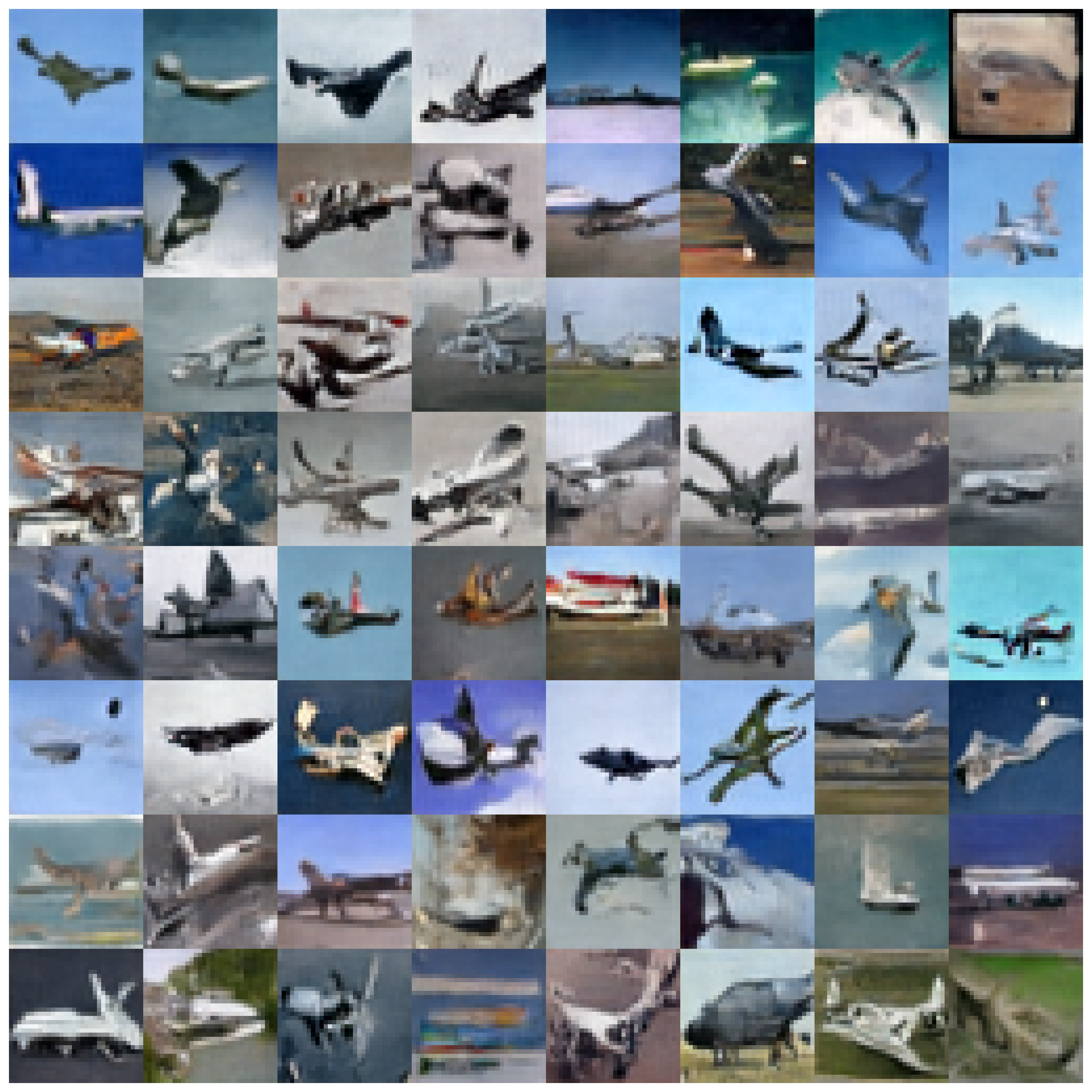}
        \caption{class: airplane}
        \label{fig:cifar-airplane}
    \end{subfigure}
    \begin{subfigure}[b]{0.33\textwidth}
        \centering
        \includegraphics[width=\linewidth]{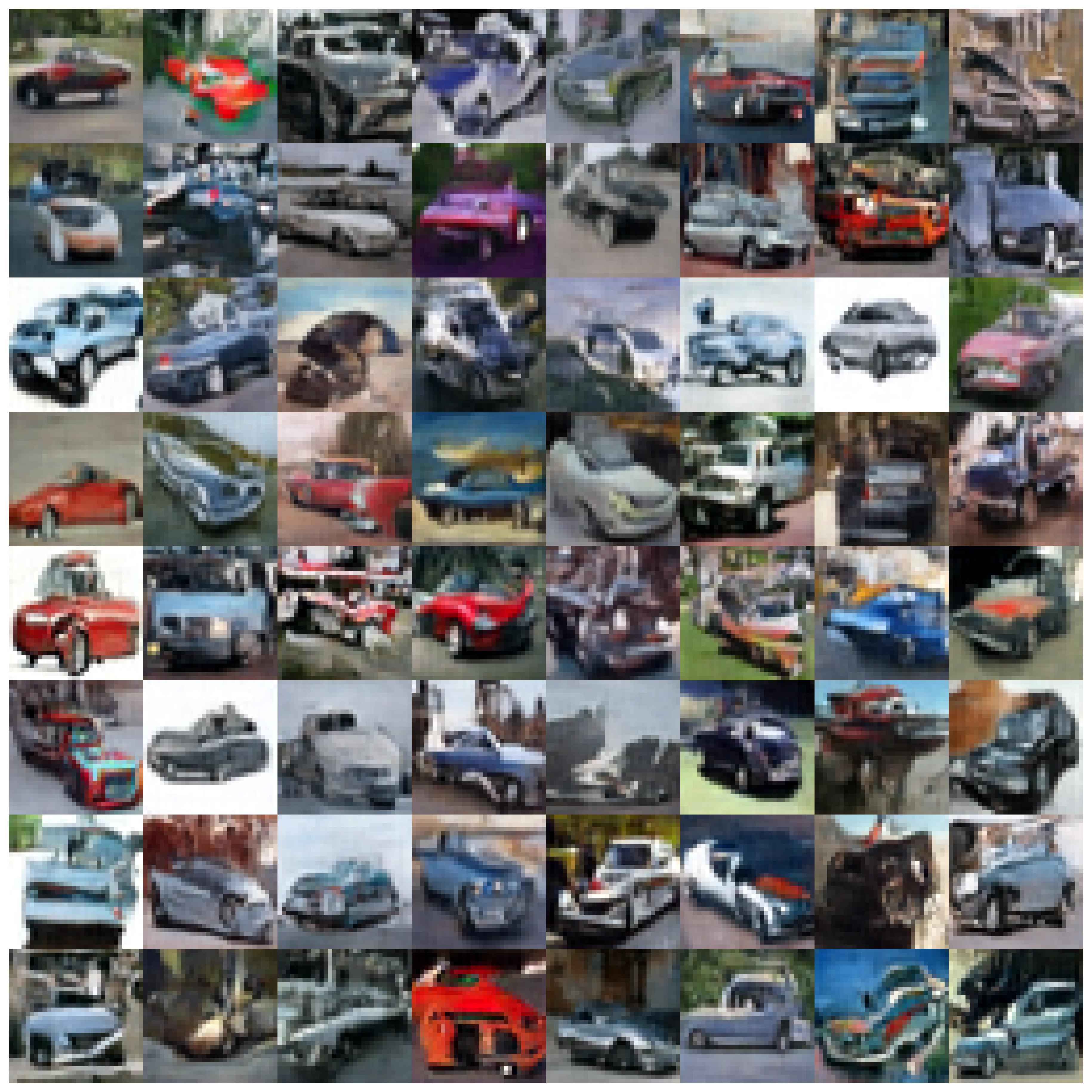}
        \caption{class: automobile}
        \label{fig:cifar-automobile}
    \end{subfigure}
    \begin{subfigure}[b]{0.33\textwidth}
        \centering
        \includegraphics[width=\linewidth]{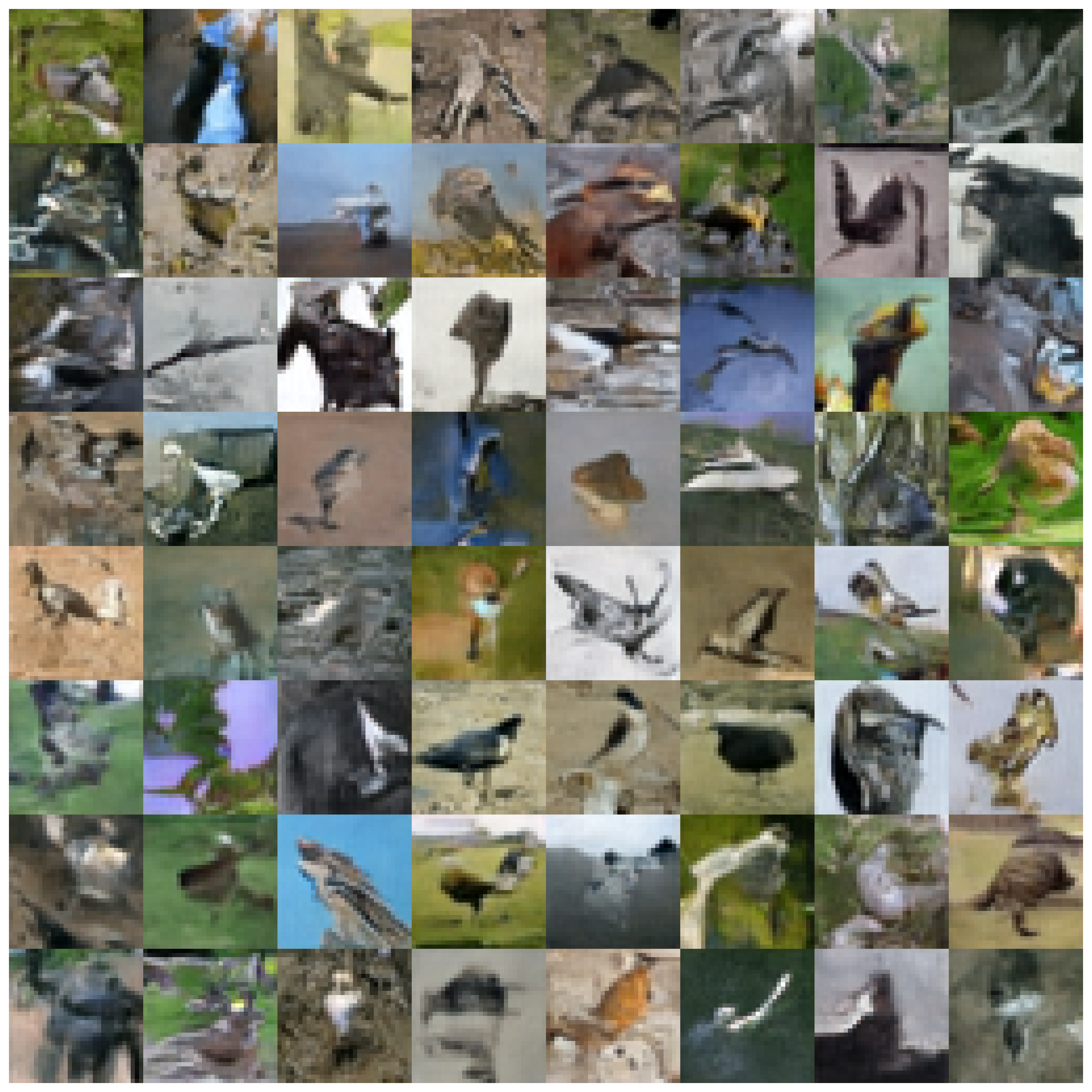}
        \caption{class: bird}
        \label{fig:cifar-bird}
    \end{subfigure}

    \vspace{0.08em}

    \begin{subfigure}[b]{0.33\textwidth}
        \centering
        \includegraphics[width=\linewidth]{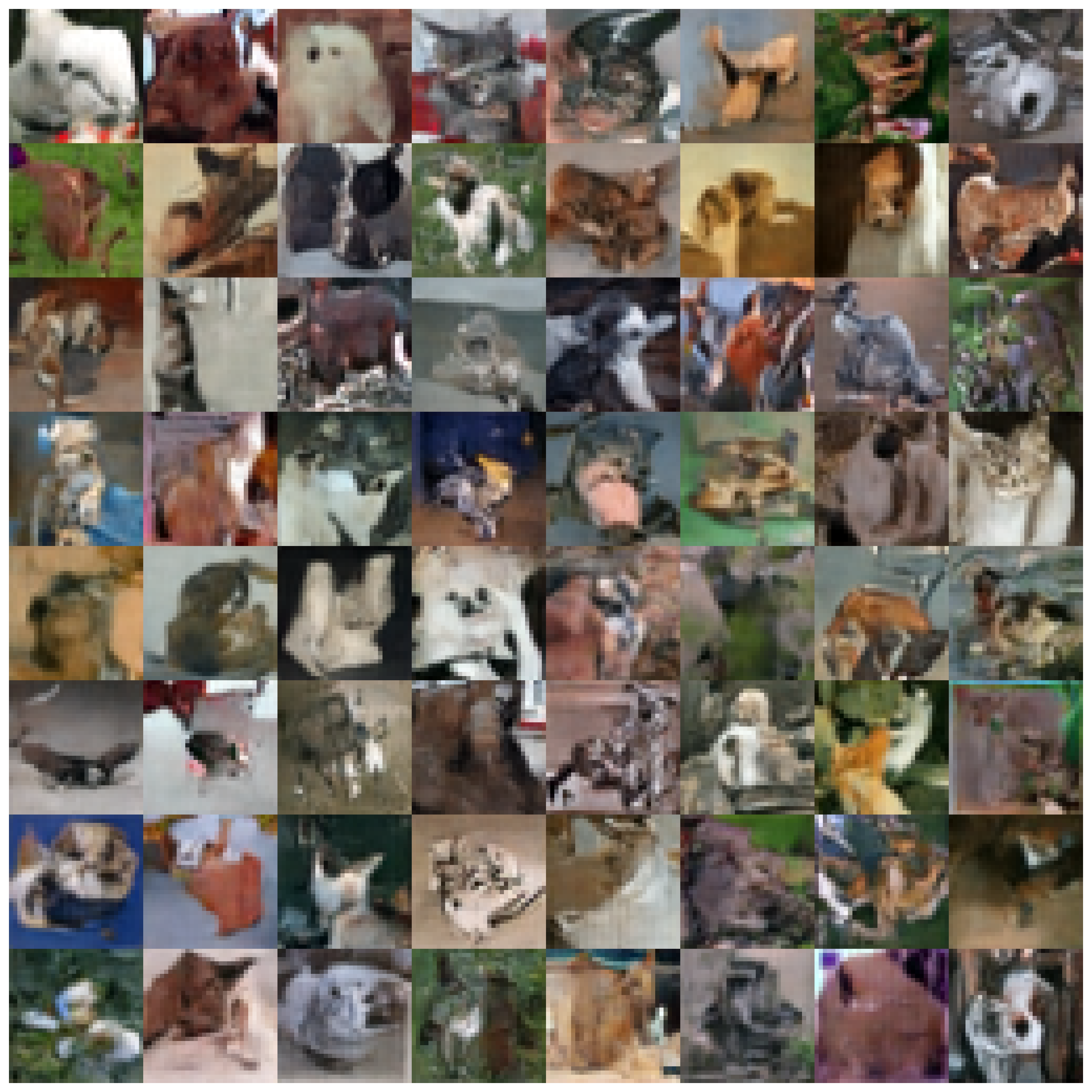}
        \caption{class: cat}
        \label{fig:cifar-cat}
    \end{subfigure}
    \begin{subfigure}[b]{0.33\textwidth}
        \centering
        \includegraphics[width=\linewidth]{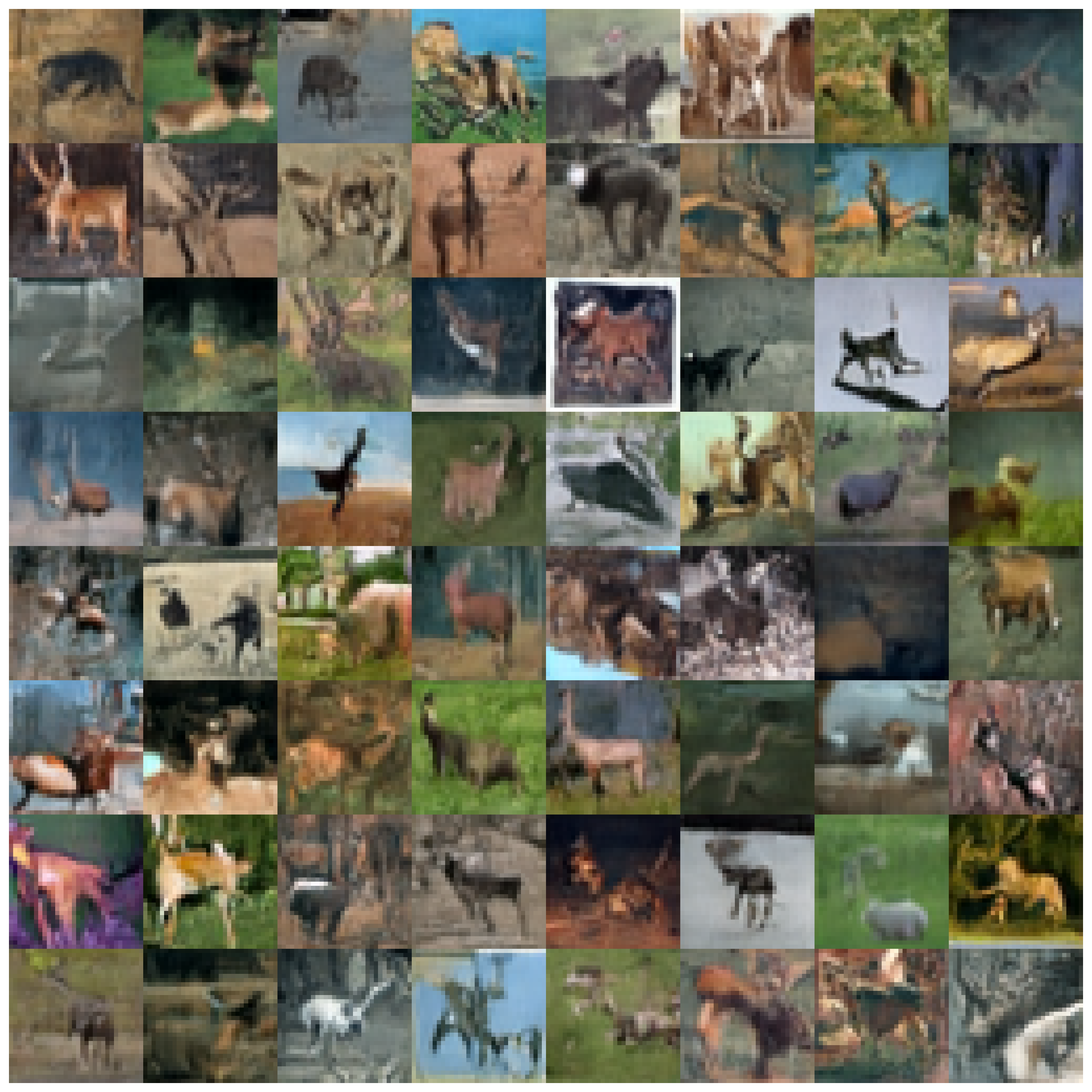}
        \caption{class: deer}
        \label{fig:cifar-deer}
    \end{subfigure}
    \begin{subfigure}[b]{0.33\textwidth}
        \centering
        \includegraphics[width=\linewidth]{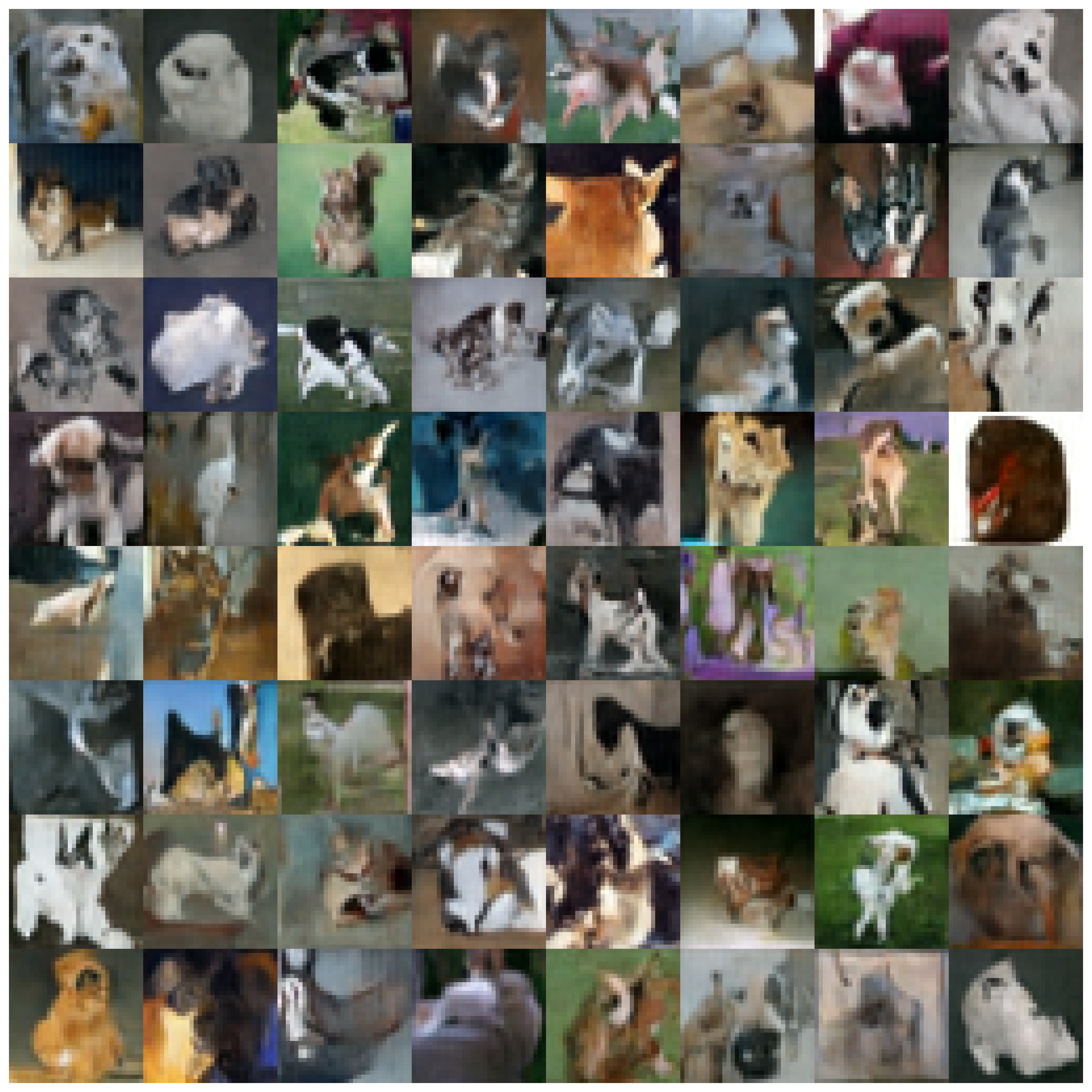}
        \caption{class: dog}
        \label{fig:cifar-dog}
    \end{subfigure}

    \vspace{0.08em}

    \begin{subfigure}[b]{0.33\textwidth}
        \centering
        \includegraphics[width=\linewidth]{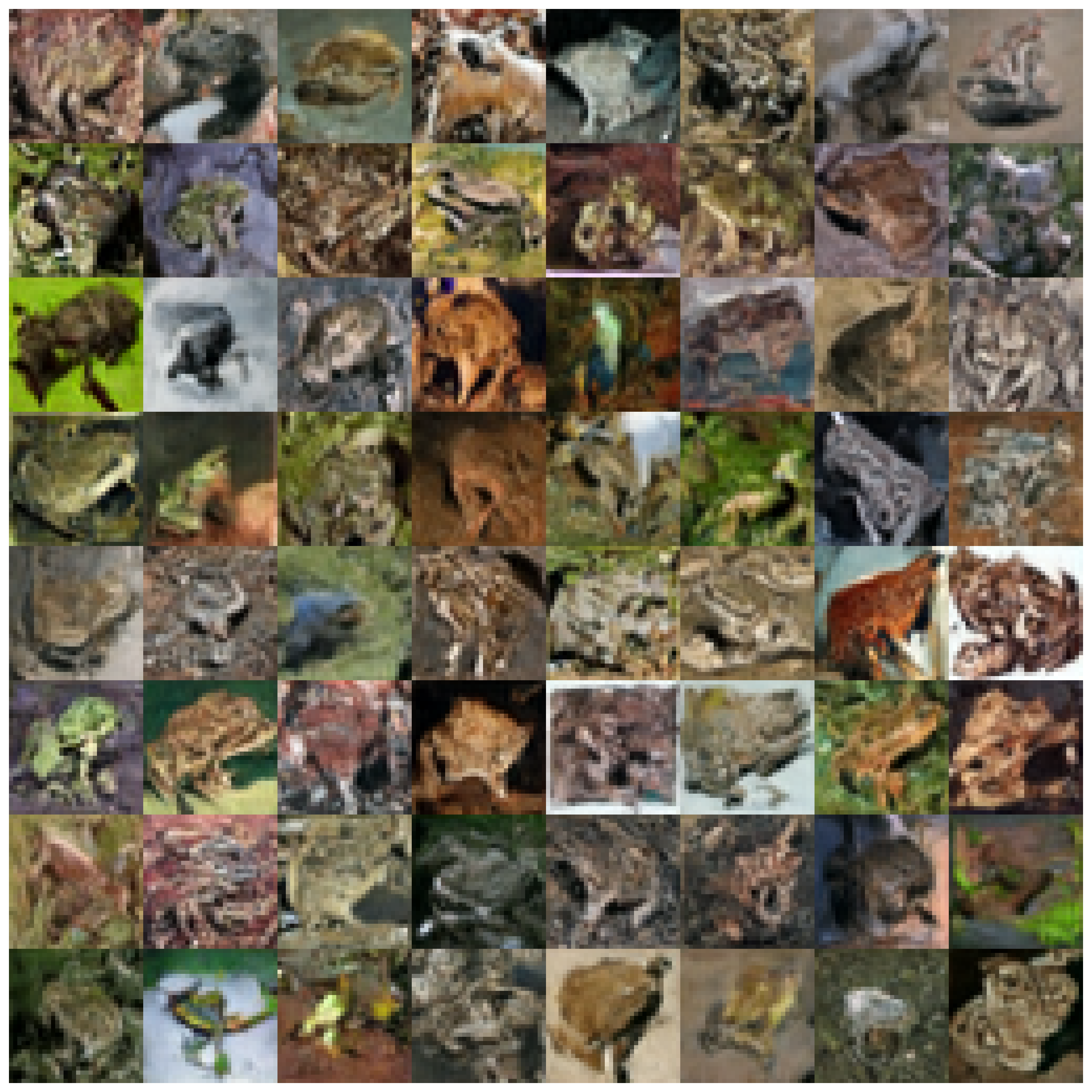}
        \caption{class: frog}
        \label{fig:cifar-frog}
    \end{subfigure}
    \begin{subfigure}[b]{0.33\textwidth}
        \centering
        \includegraphics[width=\linewidth]{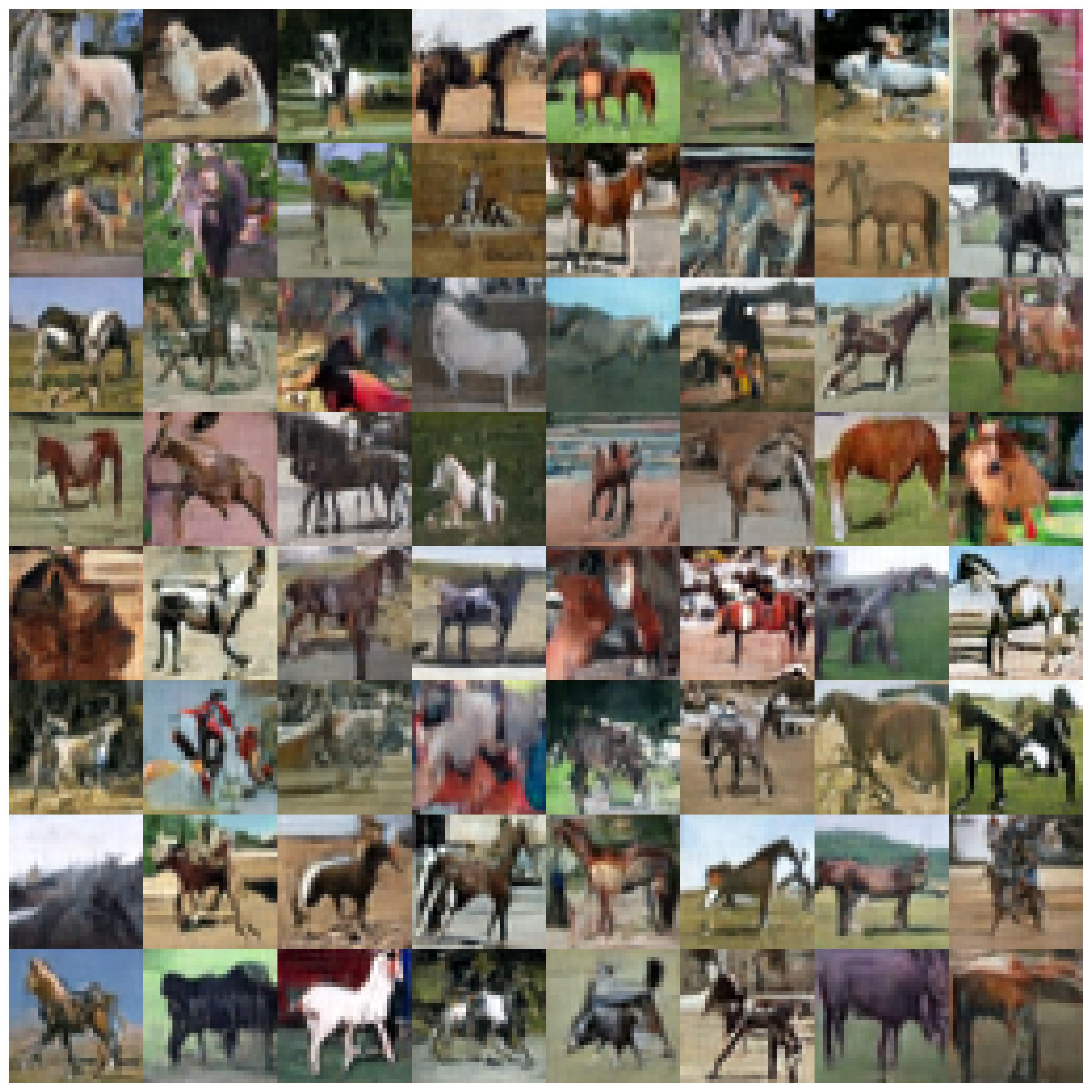}
        \caption{class: horse}
        \label{fig:cifar-horse}
    \end{subfigure}
    \begin{subfigure}[b]{0.33\textwidth}
        \centering
        \includegraphics[width=\linewidth]{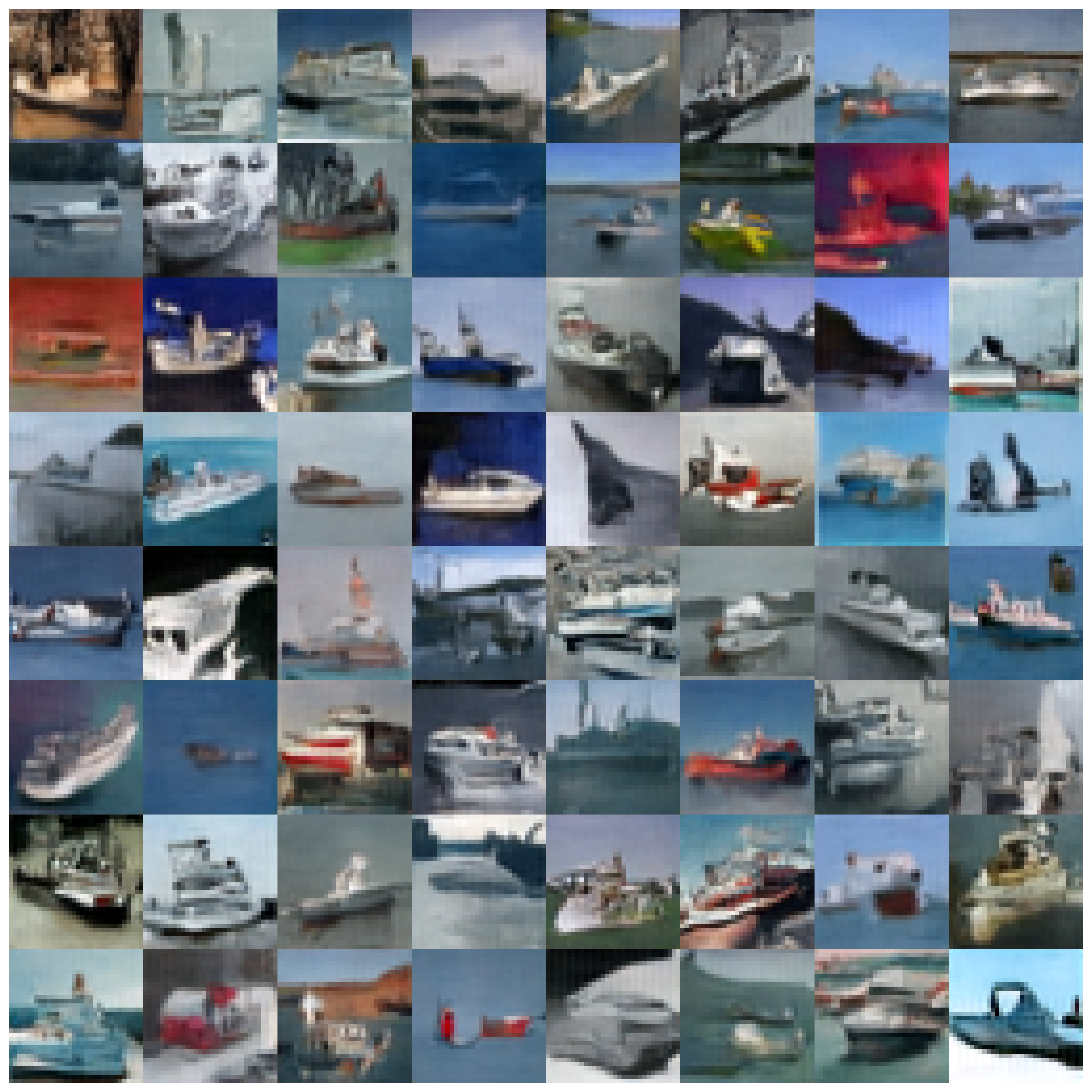}
        \caption{class: ship}
        \label{fig:cifar-ship}
    \end{subfigure}



    \caption{Class-conditional image generation results for each CIFAR-10 ($32\times 32$) class demonstrate that PreSS consistently produces images that are visually and statistically consistent with their respective specified categories.}
    \label{fig:cifar-10-subplots}
\end{figure*}

Class-conditional samples are shown in Figure~\ref{fig:results_cifar10}, and an extended set of examples is provided in Figure~\ref{fig:cifar-10-subplots}. We perform class-conditional generation on the CIFAR-10 dataset, where the sampling process starts from random noises with corresponding class-related mean and variance. Examples highlight that class-related spectral features can be preserved during our diffusion process.

\section{Implementation Details}
For CIFAR-10 experiments, both improved DDPM and our PreSS use a UNet model with $[C,2C,2C,2C]$ channels with $C=128$. For CelebA experiments, both of them use a UNet model with $[C,2C,3C,4C]$ channels with $C=128$. 

For both experiments, we use a batch size of 128, a learning rate of $10^{-4}$, an exponential moving average (EMA) over model parameters with a rate of $0.9999$, a dropout rate of $0.3$, residual blocks of three, and iteration steps of $500K$. 

For CelebA-HQ experiments, we use a pretrained VQ-AVE encoder ($f=4,\ Z=8192,\ d=3$) for both LDM and PreSS. During the training process, PreSS uses the same configuration as a pretrained LDM model (LDM-VQ-4). 

\end{document}